\documentclass[journal]{IEEEtran}
\usepackage{amsmath,amsfonts}
\usepackage{algorithmic}
\usepackage{algorithm}
\usepackage{array}
\usepackage[caption=false,font=normalsize,labelfont=sf,textfont=sf]{subfig}
\usepackage{arydshln}
\usepackage{textcomp}
\usepackage{stfloats}
\usepackage{multirow}
\usepackage{amssymb}
\usepackage{url}
\usepackage{verbatim}
\usepackage{amsmath}
\usepackage{graphicx}
\usepackage{cite}
\usepackage{xcolor}
\usepackage{color,soul}
\usepackage[colorlinks=true, allcolors=blue]{hyperref}
\newbox{\myorcidaffilbox}
\sbox{\myorcidaffilbox}{\large\includegraphics[height=4mm]{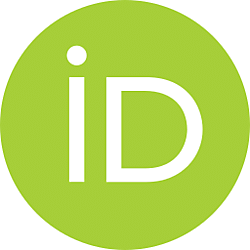}}
\newcommand{\orcidaffil}[1]{\href{https://orcid.org/#1}{\usebox{\myorcidaffilbox}}}

\newcommand\numberthis{\addtocounter{equation}{1}\tag{\theequation}}

\begin{document}

\title{PIP-Net: Pedestrian Intention Prediction in the Wild}

\author{Mohsen Azarmi \orcidaffil{0000-0003-0737-9204}, \and Mahdi Rezaei \orcidaffil{0000-0003-3892-421X}, \and He Wang \orcidaffil{0000-0002-2281-5679}

\thanks{Received 19 February 2024; revised 18 March 2025; accepted 5 May 2025. This work was supported by European Union's Horizon 2020 Research and Innovation Programme under Grant 101006664. The associate Editor for this article was V. Chamola. (\textit{Corresponding author: Mohsen Azarmi)}}
\thanks{Mohsen Azarmi and Mahdi Rezaei are with the Institute for Transport Studies, Computer Vision and Machine Learning Group, University~of~Leeds, LS2 9JT Leeds, U.K. (e-mail: {\tt tsmaz@leeds.ac.uk} and {\tt  m.rezaei@leeds.ac.uk}) }%
\thanks{He Wang is with UCL Centre for Artificial Intelligence, Department of Computer Science, University College London, WC1E 6BT London, U.K. (e-mail: {\tt he\_wang@ucl.ac.uk})}%
\thanks{Digital Object Identifier 1-.1109/TITS.2025.3570794} 

}



\maketitle
\begin{abstract}
Accurate pedestrian intention prediction (PIP) by Autonomous Vehicles (AVs) is one of the current research challenges in this field. In this article, we introduce PIP-Net, a novel framework designed to predict pedestrian crossing intentions by AVs in real-world urban scenarios. 
We offer two variants of PIP-Net designed for different camera mounts and setups. Leveraging both kinematic data and spatial features from the driving scene, the proposed model employs a recurrent and temporal attention-based solution, outperforming state-of-the-art performance. 
To enhance the visual representation of road users and their proximity to the ego vehicle, we introduce a categorical depth feature map, combined with a local motion flow feature, providing rich insights into the scene dynamics. 
Additionally, we explore the impact of expanding the camera's field of view, from one to three cameras surrounding the ego vehicle, leading to an enhancement in the model's contextual perception. Depending on the traffic scenario and road environment, the model excels in predicting pedestrian crossing intentions up to 4 seconds in advance, which is a breakthrough in current research studies in pedestrian intention prediction.
Finally, for the first time, we present the Urban-PIP dataset, a customised pedestrian intention prediction dataset, with multi-camera annotations in real-world automated driving scenarios.
\end{abstract}


\section{Introduction}
\IEEEPARstart{P}{edestrians} are the most vulnerable road users and face a high risk of fatal accidents \cite{yannis2020vulnerable}. Ensuring pedestrian safety in automated driving, particularly in mixed AV-pedestrian traffic scenarios, heavily relies on the AV's capability in ``pedestrian intention prediction (PIP)" \cite{sharma2022pedestrian}. 
A PIP system determines if a pedestrian is likely to cross the road shortly (within the next few seconds). 
This study aims to investigate the critical visual clues that pedestrians exhibit when they intend to cross the road, and then provide a model which predicts crossing behaviour, a few seconds in advance.

Anticipating pedestrian crossing behaviour is a difficult task due to various environmental factors that affect human intention \cite{najmi2023human,zhou2023pedestrian,fang2024behavioral}. Even in the simulated scenarios in which the majority of parameters are under control, crossing prediction is a challenging endeavour \cite{kalantari2023goes}. Factors like interactions with other pedestrians, traffic signs, road congestion, and vehicle speed can influence pedestrians' tendency to cross the road in front of AVs \cite{yang2021crossing}. 

\begin{figure}[!t]
    \centering
    \includegraphics[width=3.3in]{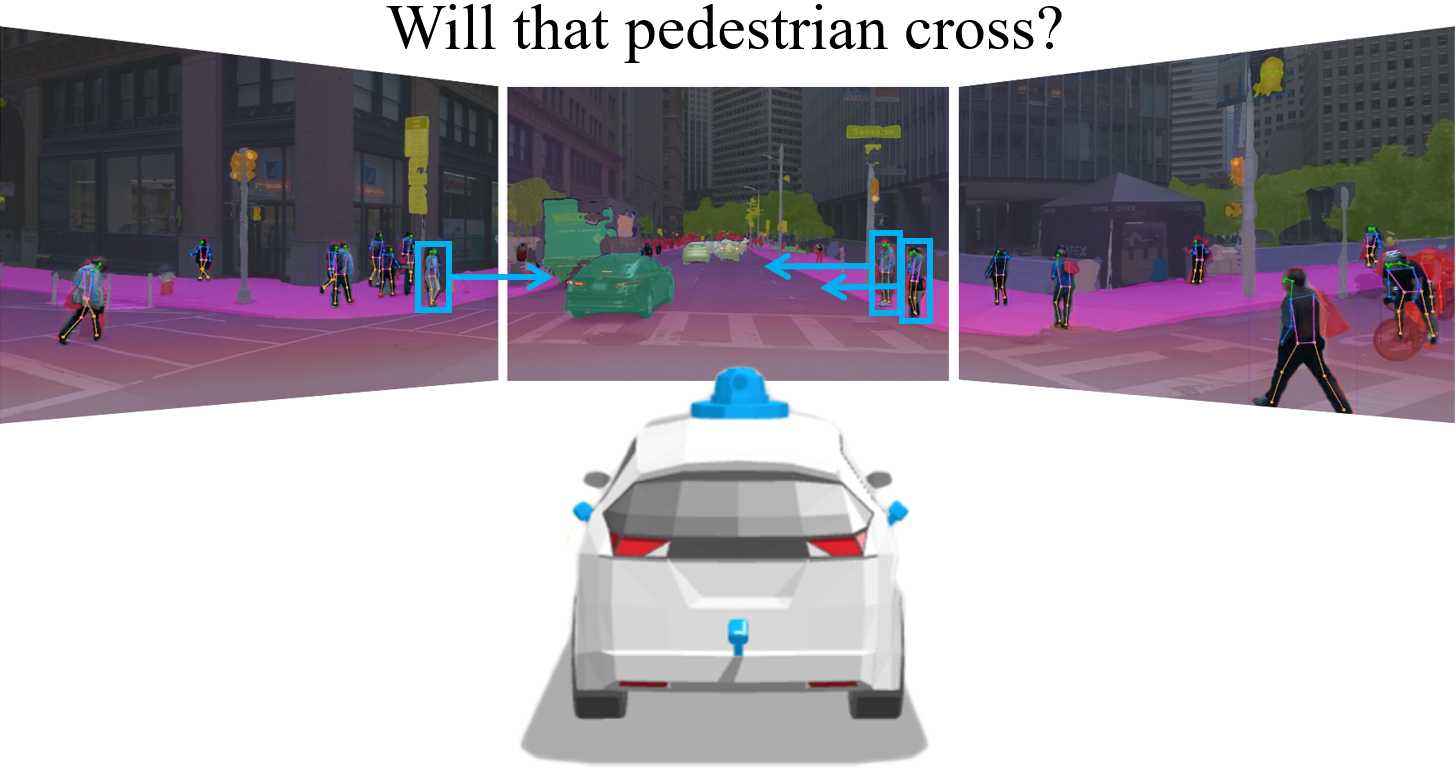}
    \vspace{-1mm}
    \caption{
    Pedestrians' crossing intention prediction in complex urban scenarios via contextual data analysis and a multi-camera perception setup. 
    }
    \label{fig:teaser}
    \vspace{-15pt}
\end{figure}

Computer vision plays a crucial role in enabling AVs to perceive their surrounding environment by analysing the visual data captured via multiple sensors, such as cameras, LiDAR, Radar, etc. Learning-based models, in particular deep neural networks (DNNs), have shown remarkable success in various computer vision tasks, including scene understanding, semantic segmentation \cite{muhammad2022vision}, road users classification, localisation \cite{chen2021deep}, and motion prediction \cite{gulzar2021survey}. Figure \ref{fig:teaser} illustrates some of the perceivable factors such as depth, pedestrian pose, and surrounding objects, that an autonomous vehicle should consider to interpret the scene and estimate the pedestrians' intention. 
DNNs are particularly effective at learning complex patterns and features from visual data, making them a natural fit for tasks that involve analysing images or videos to comprehend pedestrian behaviour \cite{ham2022mcip,ham2023cipf,ni2023pedestrians}. 
They also offer significant capabilities in multi-modal integration by providing a neural-based mechanism to process and fuse all the perceived information from diverse sensors. This integration may enhance the overall understanding of the environment and help to make more accurate and safer decisions \cite{zhang2022mutr3d}.   

Several datasets, such as JAAD \cite{rasouli2017ICCVW}, PIE \cite{rasouli2019pie}, and STIP \cite{liu2020spatiotemporal}, use onboard camera recordings and their data are publicly released for the study of pedestrians' behaviour before and during road crossing. However, most of the current research works suffer from supplying a multi-camera setup to leverage the benefits of sensor fusion and multi-modal perception.
In addition to the above-mentioned datasets, some baseline approaches \cite{kotseruba2021benchmark} have also been established for analysing the visual cues and signals that pedestrians emit through their body language and positioning. The approaches highlight the benefits of combining these features with contextual information.
Contextual information may include factors such as the road's location, the time of day, weather conditions, the presence of traffic signals or crosswalks, the type of road (urban, suburban, rural), and the position and behaviour of other vehicles near the scene \cite{schneemann2016context}. To the best of our knowledge, no extensive research has been conducted to understand and interpret such contextual details and their effects on pedestrian's decision-making.

In this study, we propose a customised DNN-based framework, called ``PIP-Net" that takes various features of pedestrians, the environment, and the ego-vehicle state into account, to learn the context of a crossing scenario and consequently predict the crossing intention of pedestrians in real-world AV urban driving scenarios. The main contributions of this research are highlighted as follows:
\begin{itemize}
    \item A novel feature fusion model is presented to integrate AV's surrounding cameras and combine visual and non-visual modalities, as well as a hybrid feature map that incorporates depth and instance semantic information of each road user to comprehend the latent dynamics in the scene. 
    \item Introducing the multi-camera ``Urban-PIP dataset'', which includes various real-world scenarios of pedestrian crossing for autonomous driving in urban areas. 
    \item We examine the effectiveness of the various input features, temporal prediction expansion, and the worthiness of expanding the vehicle field of view from one camera to three cameras based on the latest Waymo car camera setup \cite{mei2022waymo} to ensure the developed model will be in line with the current technology developments in the AV industry.
    \item Finally, we evaluate the effectiveness of the proposed model on the widely utilised PIE dataset and the introduced Urban-PIP dataset, outperforming the state-of-the-art (SOTA) for crossing prediction. 
\end{itemize}

\section{Related Works}\label{sec:related}
Recently, pedestrian crossing intention prediction research has surged and gained significant attention within the autonomous driving research community \cite{ham2022mcip,ham2023cipf,ni2023pedestrians}. 
Most current methods mainly address the problem by taking two aspects into account: Discovering influential factors and features for interpreting road users' interactions \cite{kalantari2023goes,yang2021crossing,zhou2023pedestrian}, and designing the analytic model to predict the pedestrians' crossing intention \cite{schneemann2016context,rasouli2017ICCVW,rasouli2019pie,liu2020spatiotemporal,kotseruba2021benchmark}.
Both research directions mainly utilise advanced learning-based techniques.
Deep learning methods have been fostered on multiple features of pedestrians and the environment, whether derived from annotations, visual information from videos or their combinations \cite{sharma2022pedestrian,zhang2023pedestrian}. 
The following two subsections are dedicated to introducing approaches that utilise DNN-based architectures for spatio-temporal analysis and feature selection/fusion.

\subsection{Spatio-temporal Analysis}
Recently, there has been a shift from still image analysis to the incorporation of temporal information into the prediction models. Rather than relying on individual images, most contemporary methods utilise sequences of input images for decision-making by their prediction models. This adaptation recognises the significance of temporal data in enhancing the prediction task, resulting in what is known as spatio-temporal modelling.

Spatio-temporal modelling can be achieved through a two-step process. Initially, visual (spatial) features per frame can be extracted using a visual encoder such as 2D convolutional neural networks (CNNs) \cite{razali2021pedestrian}, vision transformers (ViTs) \cite{Lorenzo2021CAPformerPC}, or graph convolution networks (GCNs) \cite{liu2020spatiotemporal}. Subsequently, these extracted features are then fed into a temporal encoder like recurrent neural networks (RNNs) \cite{lorenzo2020rnn}, using long short-term memory (LSTM) \cite{sharma2022intelligent,abughalieh2020predicting,alofi2024pedestrian} or the gated recurrent unit (GRU) modules \cite{cho2014learning, Rasouli2020PedestrianAA, bhattacharyya2018long}.
For instance, in \cite{abughalieh2020predicting,kotseruba2020they,lorenzo2020rnn}, 2D convolutions are employed to extract visual features from image sequences, while RNNs encode the temporal relationships among these features. These sequentially encoded visual features are then inputted into a fully-connected layer to generate the ultimate crossing prediction. 

An alternative approach is extracting both spatial and temporal features involves the utilisation of 3D CNNs (Conv3D) \cite{tran2015learning} or Transformer architecture \cite{vaswani2017attention}. 
3D CNNs can directly capture spatio-temporal features by substituting the 2D kernels within the convolution and pooling layers of a 2D CNN with their 3D equivalents. For instance, in works such as \cite{saleh2019real,saleh2020spatio,singh2021multi,yan2025three}, a framework based on a 3D CNN is employed to directly extract sequential visual features from sequences of pedestrian images. The ultimate prediction is then made using a fully-connected layer. Transformer architecture uses self-attention mechanisms \cite{zhao2020exploring} to capture long-range relationships, both within a single frame (spatial) and between different frames (temporal). This helps in thoroughly analysing pedestrian dynamics \cite{zhang2023trep,osman2023tamformer,achaji2022attention,zhou2023pit}. Furthermore, hybrid models combining CNNs, RNNs, and attention mechanisms have been explored \cite{yang2022predicting,sharma2023visual,azarmi2023local} to leverage the strengths of these approaches.

\subsection{Feature Selection and Integration}
Reliance on a single pedestrian feature for crossing prediction, such as using only the pedestrian's pose kinematics \cite{gesnouin2021trouspi} or bounding box \cite{achaji2022attention}, has resulted in context-free crossing predictions that neglect other scene-specific modalities and miss traffic and situational awareness.
Instead, it is possible to treat various types of information, such as the pedestrian's image, body pose keypoints, bounding box, vehicle dynamics, and the broader contextual backdrop as distinct input channels for the prediction model \cite{ni2023pedestrians}. 

Studies such as \cite{fang2018pedestrian,fang2020intention,cadena2022pedestrian} have incorporated human poses or skeletons into pedestrian crossing prediction tasks alongside various features such as pedestrian's local image, bounding box, and vehicle speed, to construct the intention classifier. This approach has shown improved prediction accuracy but often neglects other important features or lacks proper attention to feature integration.

\begin{figure}[!t]
    \centering
    \includegraphics[width=2.9in]{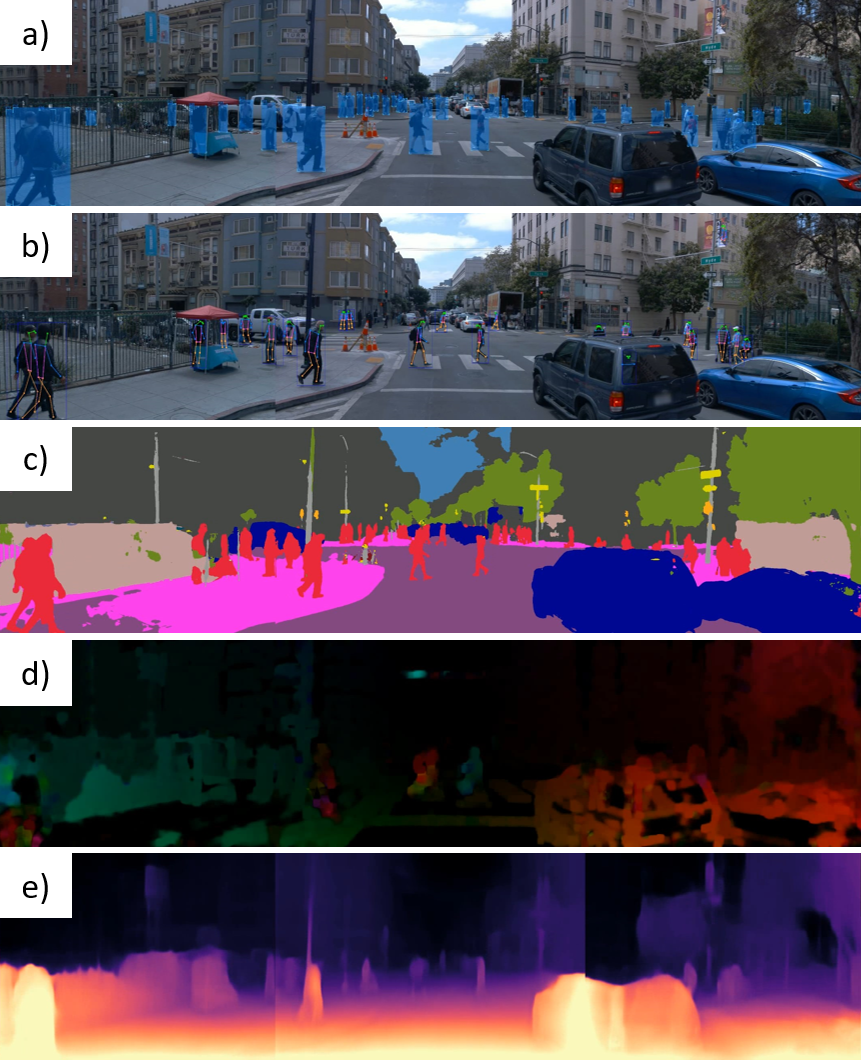}
    \vspace{-2mm}
    \caption{
    Analysing pedestrians' features and traffic scene dynamics through: 
    a) Pedestrian localisation, highlighted with bounding boxes,
    b) Gesture understanding with pose estimation,
    c) Object categorisation through segmentation,
    d) Global motion patterns using optical flow, and
    e) Estimating distance via a global depth heatmap.}
\label{fig:features}
    \vspace{-15pt}
\end{figure}

The investigation into the types of features, such as pedestrians and environmental context, is still ongoing. For instance, pedestrian-to-vehicle distance is considered one of the most influential factors in pedestrians' decisions to cross \cite{kalantari2023goes}. This feature is typically estimated as a single measure between the target pedestrian and the ego vehicle \cite{zhou2023pedestrian,xu2024pedestrian}, while relative velocity and closing speed, which refers to how quickly the gap between the ego vehicle and the pedestrian is decreasing, are also investigated \cite{azarmi2024feature}. Alternatively, a depth map of the scene (see Figure \ref{fig:features}e) can be used to assess the distance from other road users and possibly reveal the underlying dynamics \cite{neogi2020context}. However, the depth map is susceptible to noise due to rough estimation, which can lead to inaccuracies in scenarios involving multiple pedestrians crossing \cite{zhang2021pedestrian}.

On the other hand, some studies specifically concentrated on feature integration. For instance, vision and non-vision branches fusion \cite{yang2022predicting} suggest how to efficiently combine diverse data modalities at different stages of a DNN model to surge the intention prediction accuracy. Another study  \cite{ham2022mcip} is conducted to merge two visual and three non-visual elements of the pedestrian, scene, and subject vehicle in a multi-stream network. From a different perspective, in studies such as \cite{ham2023cipf,azarmi2023local}, local and global contextual information has been weighted by an attention mechanism and fused together to apply a prediction on \textit{Joint Attention in Autonomous Driving} (JAAD) \cite{rasouli2017ICCVW} and \textit{Pedestrian Intention Estimation} (PIE) \cite{rasouli2019pie} datasets.

\subsection{Research Gaps}
Despite advancements in pedestrian crossing intention prediction, several key gaps persist, as outlined in recent surveys \cite{sharma2022pedestrian,zhang2023pedestrian,fang2024behavioral}. In this study, we focus on addressing four research gaps:

1) Pedestrian crossing intention highly relies on the distance of the AV to the pedestrian and the relative distance of the pedestrian to other road users, which may fall into various categories of instance segmentation (e.g., cars, other pedestrians, etc.). None of the reviewed research has considered the simultaneous impact of both features on pedestrian intention. 

2) To the best of our knowledge, no prior study has considered instance segmentation to smooth and normalise the distance measurement of road user instances. 
In this article, we propose a new concept of \textit{Categorical Depth} which integrates the classic noisy depth measurement with instance segmentation to gain more accurate depth information and provide context-aware spatial data.

3) The reviewed models, often have limited generalisability and are incapable of performing in the wild and real-world automated driving scenarios, as they have normally been tested with a human-driven vehicle
\cite{gesnouin2022assessing}.
Our study focuses on the real-world Waymo dataset, which is collected from an AV's field of view. Our added pedestrian intention annotations to this dataset help address the issue of inadequate annotated ground-truth data availability in this field of research.

4) There is a shortage of dedicated neural network architectures capable of effectively accommodating and extracting maximal multi-camera information from around the AV for context recognition, hence an accurate model for predicting pedestrian crossing intentions.
Camera integration is proposed in this study to cope with the limited field of view.

\section{Methodology}
We propose the PIP-Net model, which is based on deep neural networks for predicting pedestrian crossing intention. The model incorporates spatial-temporal features such as road users' positioning, pose, and dynamic movements, along with a hybrid feature map, the categorised semantic and depth information as input to the network.
A multi-camera stitching and integration model is developed to facilitate panoramic viewing, enabling a synchronised pedestrian ID assignment and tracking across the entire multi-view scene, thus enhancing the PIP-Net model's understanding of spatial characteristics and contextual information.

\begin{figure*}[!t]
    \centering
    \includegraphics[width=7in]{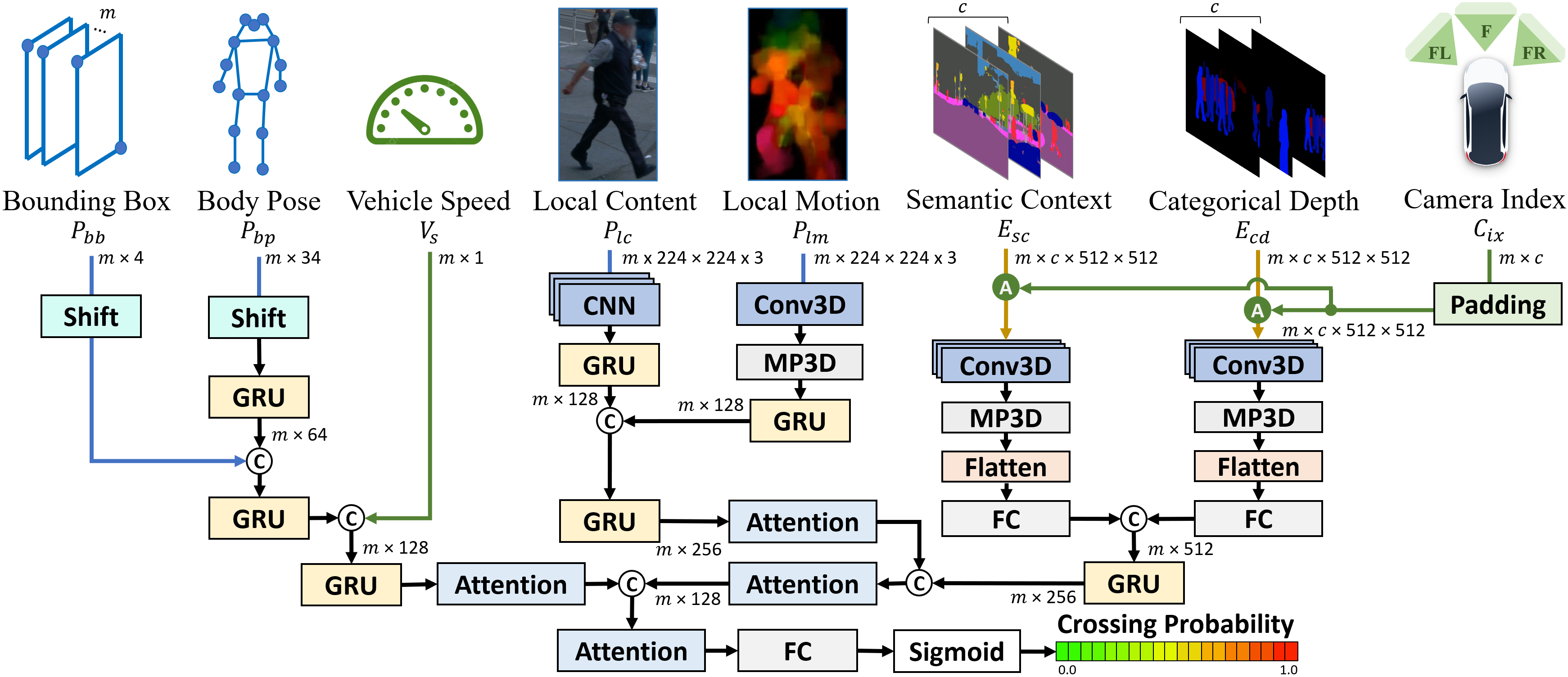}
    \caption{\textbf{The overview of the proposed DNN-based framework.} 
    PIP-Net receives \textit{Spatial Kinematic} and \textit{Spatial Context} data, and applies multi-modal feature fusion and multi-camera integration with temporal and attention-based analysis to predict the pedestrians' crossing intentions. 
    Shift and Padding units adjust the location of the target pedestrian with respect to the corresponding camera. Aggregation module (A) fuses the cameras' features as shown in Figure \ref{fig:aggregation}.
    }
    \label{fig:method}
    \vspace{-15pt}
\end{figure*}

An overview of the proposed architecture is illustrated in Figure \ref{fig:method}. The input features are categorised into spatial kinematic data and contextual data, and they are passed to the model through distinct pipelines based on their data type.
Finally, the recurrent module and attention module are utilised to improve temporal data processing.

\subsection{Spatial Kinematics}
Kinematic input data includes the positioning of the pedestrian in the scene with reference to the detected pedestrian bounding box $P_{bb}$, pedestrian body pose keypoints $P_{bp}$, and the ego-vehicle speed $V_s$. 

The data is arranged in a gated recurrent unit (GRU) layer \cite{cho2014learning}, beginning with the \textit{Bounding Box} feature $P_{bb}$. It indicates the location of the pedestrians which is detected through the customised You-Only-Look-Once algorithm for road user detection \cite{rezaei20233d}.
This feature is defined as:
\begin{equation}
    P_{bb} = \{ b^{t-m}_i, b^{t-m+1}_i, ..., b^{t}_i \},
\end{equation}
where $b_i = [x_1, y_1, x_2, y_2] \in \mathbb{R}^4$ represents the coordinates of a pedestrian bounding box. 
It consists of the top-left $([x1, y1])$ and the bottom-right coordinates $([x2, y2])$. The dimension of the bounding box matrix $P_{bb}$ is determined as $m \times 4$, where $m$ is the observation sequence length which indicates the number of frames that are observed to predict the pedestrian crossing.
We define $t$ as the \textit{decisive moment}, 0.5 $\thicksim$ 4 seconds before the crossing event. 

The \textit{Body Pose} feature $P_{bp}$ is defined as:
\begin{equation}
    P_{bp} = \{ p^{t-m}_i, p^{t-m+1}_i, ..., p^{t}_i \},
\end{equation}
where the pose keypoints are obtained using YOLO-Pose \cite{maji2022yolo}, which estimates the pose of a person by detecting 17 keypoints joints, including the shoulders, elbows, wrists, hips, knees, ankles, eyes, ears, and nose. The keypoints are represented by a 34-dimensional vector, $p_i$, which contains the 2D coordinates of each joint for the $i$-th pedestrian at time $t$.  

The \textit{Vehicle Speed} is also defined as:
\begin{equation}
    V_s = \{ s^{t-m}_i, s^{t-m+1}_i, ..., s^{t}_i \},
\end{equation}
where $s_i$ refers to the exact speed of the ego-vehicle in \textit{km}.

\subsection{Spatial Context}
Contextual input data includes pedestrian features, such as a pedestrian-bounded image (\textit{Local Content}, $P_{lc}$) and corresponding motion flow analysis of the pedestrian (\textit{Local Motion}, $P_{lm}$), the environment features like semantic segmentation of the scene (\textit{Semantic Context, $E_{sc}$}), as well as our proposed hybrid feature map (\textit{Categorical Depth}, $E_{cd}$), which refines depth information for specific pedestrians and vehicles in the scene. 

The \textit{Local Content} feature $P_{lc}$ is defined as:
\begin{equation}
    P_{lc} = \{ lc^{t-m}_i, lc^{t-m+1}_i, ..., lc^{t}_i \},
\end{equation}
where $lc_i$ denotes the feature vector that is output by applying the CNN backbone to an RGB image. An ImageNet pre-trained VGG19 network is used as the CNN backbone, with a maximum pooling layer as suggested in \cite{yang2022predicting}. Subsequently, a GRU is applied recursively to process the temporal dimesion of the feature. The RGB image contains an individual pedestrian, cropped based on the bounding box location and subsequently warped to dimensions of $224 \times 224 \times 3$ pixels, which is the optimum spatial size in the network \cite{Lorenzo2021CAPformerPC}.

The \textit{Local Motion} feature $P_{lm}$ is derived from the dense optical flow analysis within the pedestrian-bounded image.
This analysis is more consistent than examining the entire scene, which can be affected by ego-vehicle motions.
We opt for a more advanced optical flow approach using Flownet2 \cite{ilg2017flownet}. This deep learning-based method offers improved accuracy and faster run-time performance. The \textit{Local Motion} is defined as:
\begin{equation}
    P_{lm} = \{ lm^{t-m}_i, lm^{t-m+1}_i, ..., lm^{t}_i \},
\end{equation}
where $lm_i$ is considered as the localised $i$-th pedestrian motion descriptor. 
A Conv3D layer is used to extract a feature vector of size $(m, 512)$, where $m$ represents the observation sequence length. The fueature vector is then passed through a 3D max-pooling layer (MP3D) with a kernel size of $4 \times 4$ and a GRU module. This process yields a $(m, 128)$ vector,  this vector is then inputted into a GRU layer, resulting in an $(m, 128)$ vector, suitable for concatenation with the \textit{Local Content} feature vector.

\begin{figure}[!t]
    \centering
    \subfloat[The depth heatmap of instances]{\includegraphics[width=3.35in]{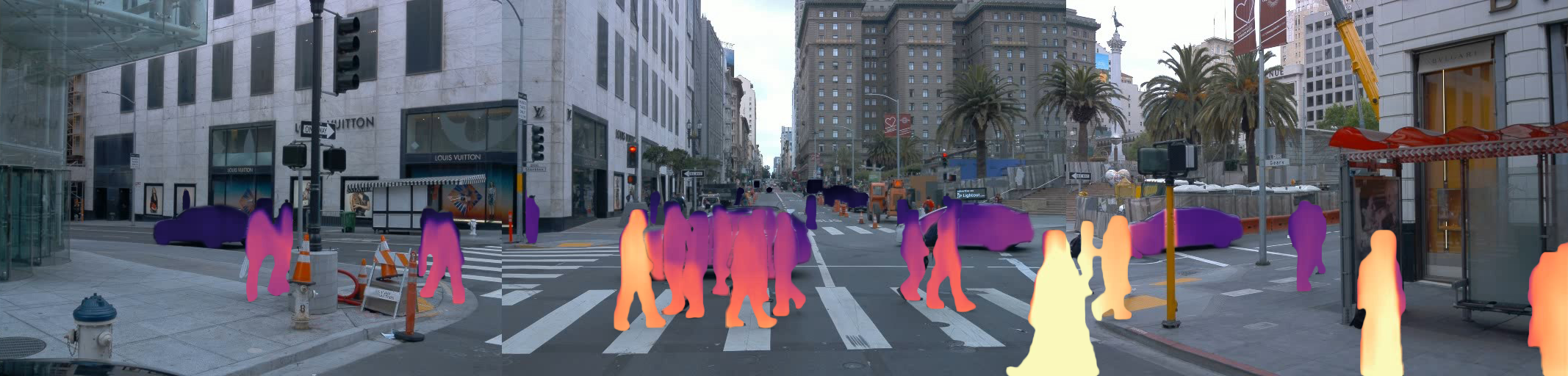}%
    \label{subfig:instance}}
    \vspace{-2mm}
    \hfil
    \subfloat[The normalised heatmap per each instance]{\includegraphics[width=3.35in]{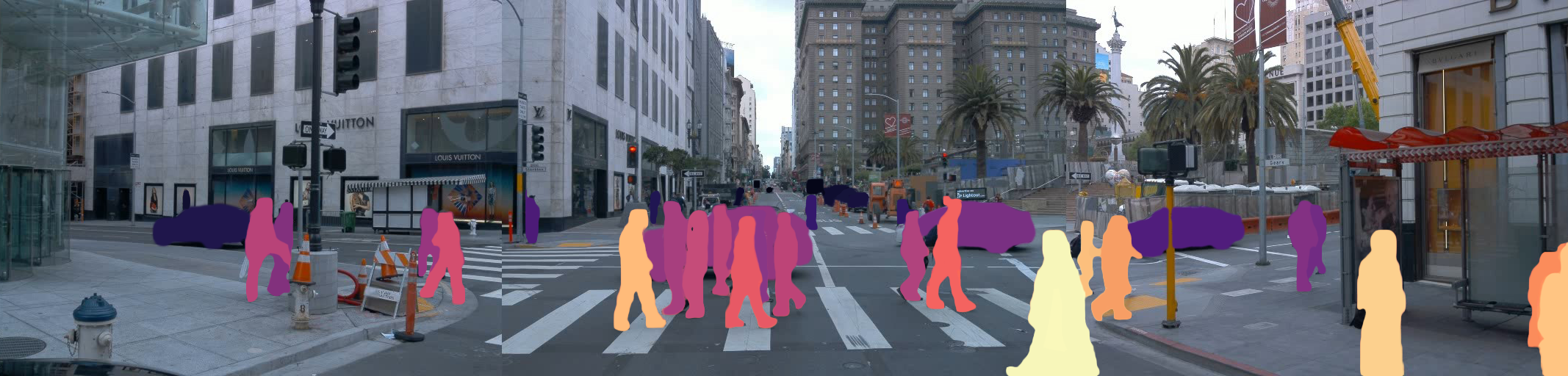}%
    \label{subfig:normal_heatmap}}
    \vspace{-2mm}
    \hfil
    \subfloat[Categorical depth map]{\includegraphics[width=3.35in]{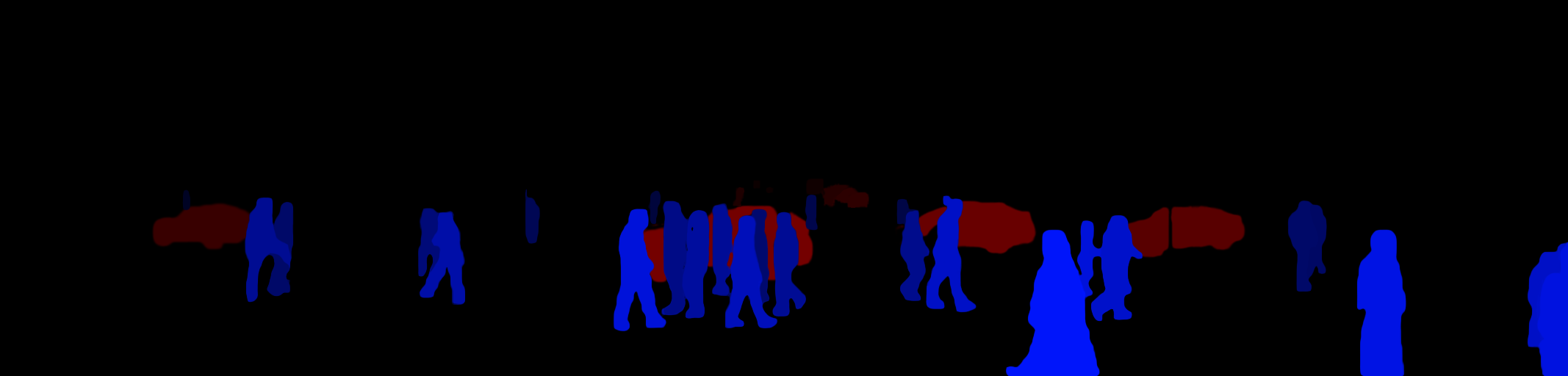}%
    \label{subfig:categorical_depth}}
    \hfil
    \caption{\textbf{Colour-coded visualisation of road users' distance to the ego-vehicle after the tripe camera stitching}. \ref{subfig:instance} is the proportion of the global depth heatmap which pedestrians and vehicles occupy; \ref{subfig:normal_heatmap} is the normalised heatmap values per each instance; \ref{subfig:categorical_depth} is a visualised $E_{cd}$ feature in an RGB format in which pedestrians and vehicles integrated into the blue and red channel, respectively, and high-intensity values indicate closer users. A categorical depth map has a positive effect on removing the camera stitching effect and increasing the clarity and saliency of pedestrians and vehicles in the scene.}
    \label{fig:categorical_depth}
    \vspace{-15pt}
\end{figure}

The \textit{Semantic Context} feature $E_{sc}$ is defined as:
\begin{equation}
    E_{sc} = \{ sc^{t-m}, sc^{t-m+1}, ..., sc^{t} \},
\end{equation}
where $sc$ refers to the semantic segmentation of objects within the entire scene encompassing road structure and users. This feature ensures that the model considers the spatial distribution of classes for both moving and static objects within the scene. The semantic information is extracted by Slot-VPS \cite{zhou2022slot} model, which is a panoptic video segmentation algorithm that not only offers semantic segments but also a unique ID for each instance of the objects in the scene. 
The segmented classes include 8 dynamic classes (person, rider, car, truck, bus, train, motorcycle, and bicycle) and 11 static classes (traffic light, fire hydrant, stop sign,  parking meter, bench, handbag, road, sidewalk, sky, building, and vegetation).
The instance segments corresponding to pedestrians and vehicles are integrated into the \textit{Categorical Depth} feature, denoted as $E_{cd}$, using the following formulation:
\begin{equation}
    E_{cd} = \{ cd^{t-m}, cd^{t-m+1}, ..., cd^{t} \},
\end{equation}
where $cd$ represents the hybrid feature map containing depth information for pedestrians and vehicles within the scene. The depth data are initially estimated using the ManyDepth model \cite{watson2021temporal} and encoded in a heatmap representation, resulting in a global depth heatmap. As illustrated in Figure \ref{fig:features}e, high-intensity spots (white and oranges) indicate proximity to the ego vehicle, while low-intensity spots (navy blue and black) represent greater distances.
However, our experiments revealed that the global depth heatmap is unreliable due to inconsistencies in providing clear object boundaries. To address this, the pedestrian and vehicle instances are cropped using instance masks obtained from the Slot-VPS, as shown in Figure \ref{subfig:instance}. Subsequently, the intensity of pixels within each instance is normalised by averaging, yielding a normalised heatmap as seen in Figure \ref{subfig:normal_heatmap}. This process ensures a clear and consistent depth estimation for each class, as depicted in Figure \ref{subfig:categorical_depth}.

Both inputs, $E_{sc}$ and $E_{cd}$, undergo extraction via three Conv3D layers to be assessed for the spatio-temporal analysis. The feature dimensions are gradually reduced $(512 \rightarrow 256 \rightarrow  128\rightarrow 64)$ by repeatedly subjecting them to max-pooling layers. 
This process not only selects the most important information from the local neighbourhood of each pooling window but also reduces the spatial dimensions (width and height) of the feature maps and the computational complexity of the network.
Then the features are organised using a flattened layer, resulting in a one-dimensional array that is suitable for concatenation and can be fed into a fully-connected layer (FC). Finally, the data is passed through a GRU module.
The outputs of the three GRUs are combined and concatenated into a single output, which is then passed through an attention mechanism.

\subsection{Cameras Features Integration} \label{camfuse}
The incorporation of multiple cameras might be beneficial for capturing complex traffic scenarios, such as intersections, thanks to providing a surrounding field of view. In these scenarios, pedestrians may approach the road from the sides rather than directly in front of the vehicle. They may also choose to cross the road while a vehicle is changing lanes or making a turn. By incorporating left and right-side cameras, we can gather critical information about pedestrians in adjacent lanes or at the side of the vehicle.
The Waymo dataset is one of the best options with three cameras ($c = 3$) that also offer a diversity of real-world pedestrian crossing scenarios. The cameras are named front-left (FL), front (F), and front-right (FR) positioned from the AV's left to right angles (as shown in Figure \ref{fig:features}). 
The synchronised videos provided have an approximately 11\% overlap along the edges. 
These overlapping areas can introduce redundancy in data and make it challenging to precisely determine the pedestrian's position, movement, and intention when it moves from one camera scene to another. 
Therefore, we merge the cameras using the Panoptic stitching over time approach \cite{mei2022waymo}, excluding the overlapping regions and giving higher priority to the front-view camera.
Figure \ref{fig:features} illustrates an example of different types of features that have been extracted from three cameras, and then stitched together to constitute a single wide image.

\begin{figure}[!t]
    \centering
    \includegraphics[width=2.6in]{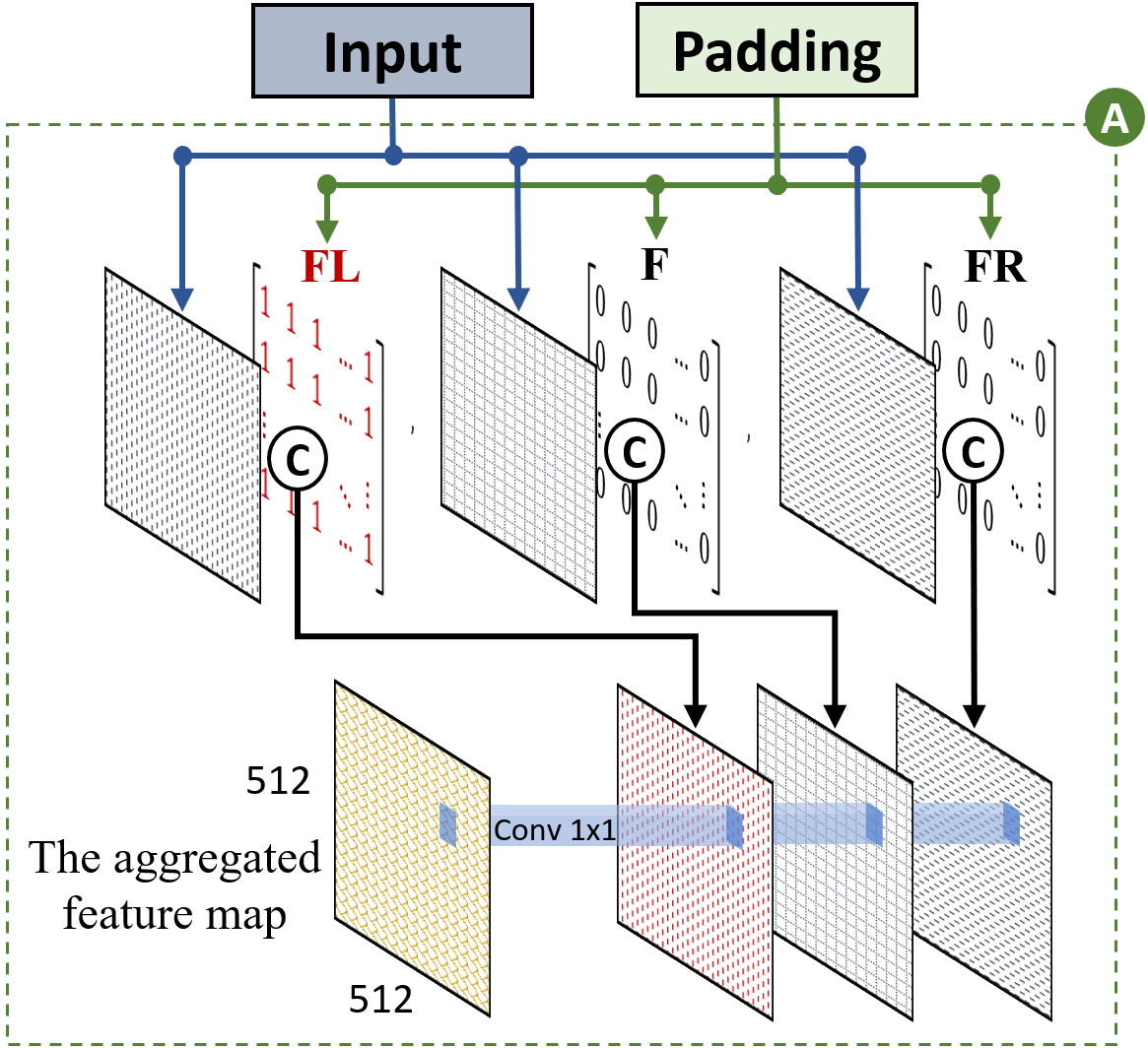}
    \vspace{-3mm}
    \caption{\textbf{The multi-camera feature aggregation module}. In this example, the front-left camera (FL) is the sentinel camera, where the target pedestrian is observed. The padding module expands the \textit{Camera Index} value and provides a camera indicator mask for combining with \textit{Sense Context} and \textit{Categorical Depth} features.}
    \label{fig:aggregation}
    \vspace{-15pt}
\end{figure}

We define \textit{Sentinel Camera} as a variable that indicates the index of the camera ($C_{ix}$) on which the target pedestrian has been observed. Using the \textit{Camera Index}, we can adjust the pose and bounding box coordinates with respect to the sentinel camera. This task will be accomplished by the Shift unit, as shown in Figure \ref{fig:method} in cyan, which extends the global coordinates from the leftmost camera to the rightmost one, and applies these adjustments to the inputs.
In this context, the Padding unit is responsible for generating a zero binary mask of size $c \times 512 \times 512$, where the $c$-th dimension corresponds to the sentinel camera being set to one. Here, $c$ represents the number of cameras. As the aggregation module (A) shown in Figure \ref{fig:aggregation}, the binary masks are combined with the input vectors i.e., $E_{sc}$ and $E_{cd}$. Subsequently, a pointwise convolution (Conv $1 \times 1$) operator aggregates all the features across cameras. This process combines features from different channels (cameras) at each spatial location, allowing a weighted combination of input features.

\subsection{Temporal and Attention Module}
To account for the temporal context of input features, GRU is employed. When describing the recursion for the GRU equation, the variables at the $j$-th level of the stack can be outlined as follows:
\begin{align*}
     z^t_j = \sigma (x^t_jW^{xz}_j + h^t_jW^{hz}_j + b^z_j), \numberthis \label{eq1}\\
     r^t_j = \sigma (x^t_jW^{xr}_j + h^t_jW^{hr}_j + b^r_j), \numberthis \label{eq2}\\
     \tilde{h}^t_j = tanh(x^t_jW^{x}_j + (r^t_j \odot r^{t-1}_j) W^h_j + b), \numberthis \label{eq3}\\
     h^t_j = (1 - z^t_j) \odot h^{t-1}_j + z^t_j \odot \tilde{h}^t_j, \numberthis \label{eq4}
\end{align*}
where $\sigma(\cdot)$ denotes the logistic sigmoid function $x^t_j$ is the input feature at time step $t$. The reset and update gates at time step $t$ are denoted as $r^t_j$ and $z^t_j$, respectively, and the weights between the two units are represented by $W$. The hidden state at the previous time step and the current time step are represented by $h^{t-1}_j$  and $h^t_j$, respectively.

To assess the significance of the processed features during network training, the attention mechanism \cite{zhao2020exploring}) is utilised to focus on specific segments of features, thereby enhancing the effectiveness of feature analysis. The resulting vector from the attention module is defined as follows:
\begin{align*}
     A = tanh(W_c [h_c : h_m]), \numberthis \label{eq5}\\
     h_c = \sum_{s^t} \alpha_t h_{s^t}, \numberthis \label{eq6}
\end{align*}
where $W_c$ represents a weight matrix, $h_c$ denotes the cumulative sum of all attention-weighted hidden states,  $h_m$ signifies the final hidden state of the encoder, $h_{s^t}$ corresponds to the preceding hidden state of the encoder, and $\alpha^t$ denotes the vector of attention weights, which is defined as follows:
\begin{align*}
     \alpha_t = \frac{\mbox{exp}(score(h_m, \tilde{h}_s)}
     {\Sigma^{t_m}_{s^t=1} \mbox{exp}(score(h_m, \tilde{h}_{s^t}))}, \numberthis \label{eq7}\\
     score(h_m, h_{s^t}) = h^T_m W_p h_{s^t} \numberthis \label{eq8}
\end{align*}
where $t_m$ is the input sequence length at time $t$. $h^T_m$ represents the transpose of the $h_m$ vector, and $W_p$ is a weight matrix that can be estimated through the training phase of the network.

In the tile of the network, the outputs of the attention modules are concatenated and then forwarded through the last attention module and an FC layer. The ultimate output, normalised to a range between zero and one using the Sigmoid function, represents the predicted probability of a pedestrian crossing.

\section{Experiments}

In this section, we conduct four distinct experiments to thoroughly assess the robustness of the proposed framework. Each experiment is designed to provide unique insights into various aspects of the model's performance. 
We compare our model against the PIE dataset in four different intervals ranging from 1 to 4 seconds, allowing us to scrutinise the model's predictive capabilities for an in-time response in different driving scenarios.
Additionally, we evaluate the model's performance under varying temporal resolution and observation time.
The second experiment examines the impact of the introduced Categorical Depth and Local Motion features in enhancing the framework's prediction accuracy.
The third experiment evaluates the model generalisability and reliability on a diverse dataset from Waymo's self-driving vehicles. This investigation ensures that the framework performs effectively across different real-world scenarios.
Lastly, we investigate the scalability of the model by assessing the framework's ability to handle one to three cameras simultaneously, expanding its view angles. This provides insights into the model's efficiency when processing information from multiple cameras.
This multi-faceted evaluation allows us to gain a detailed understanding of the model's effectiveness in comparison to a singular experiment.

\subsection{Datasets}
The JAAD \cite{rasouli2017ICCVW} and STIP \cite{liu2020spatiotemporal} datasets suffer from no annotation in terms of ego-vehicle speed values. Furthermore, there is a slight bias in these datasets, as the majority of annotations indicate the cases of crossing which may not result in effective training of deep models.
Therefore, the evaluations were conducted on the \textit{Pedestrian Intention Estimation} (PIE) \cite{rasouli2019pie} dataset, which is extensively employed in the majority of prior studies. 
Additionally, we utilised our custom dataset named Urban-PIP, specifically annotated for pedestrian crossing behaviour, built upon the Waymo \cite{mei2022waymo} dataset. Waymo is a widely used dataset for traffic perception by AVs thanks to the diversity of the video data from urban and rural environments under various driving conditions and situations.
The specifications of datasets are briefly mentioned in Table \ref{table:datasets}.

The evaluation metrics are based on \cite{kotseruba2021benchmark}, including \textit{accuracy} (Acc), \textit{precision}, and \textit{recall rate}, which quantify the model's ability to accurately predict the binary classification task. Additionally, the \textit{area under the ROC curve} (AUC), indicates the model's proficiency in distinguishing between different classes, and \textit{F1 score}, represents the harmonic mean of precision and recall rate.

\subsubsection{\bf{PIE dataset}}
The dataset is recorded on a sunny clear day for 6 hours in HD format ($1920 \times 1080$). Each video segment lasts approximately 10 minutes, resulting in a total of 6 sets. 
We utilised approximately 50\% (880 samples) of the dataset for training, 40\% (719) for testing, and 10\% (243) for validation as per the same split proportion as \cite{ham2023cipf}.
Regarding occlusion levels, partial occlusion is defined when an object is obstructed between 25\% and 75\%, while full occlusion occurs when the object is obstructed by 75\% or more.
The dataset includes the vehicle speed, heading direction, and GPS coordinates.

\begin{table}[tb]
\setlength{\tabcolsep}{4pt} 
    \centering
    \caption{Dataset Specifications}
    \vspace{-2mm}
    \begin{tabular}{lll}
        \noalign{\hrule height 1pt}
        \textbf{Specification}            & \bf{PIE}    & \bf{Urban-PIP} \\
        \hline
        Autonomous Driving                & No              & Yes      \\
        Number of Cameras                 & 1               & 3        \\
        Auxiliary Sensors                 & OBD             & LiDAR, Radar, IMU  \\
        Lighting Conditions               & Daylight        & Daylight, Nighttime \\
        Demographic Information           & Age, Gender     & None \\
        Video clip lengths                & 10 min          & 16 sec   \\
        Total Number of Frames            & 909,000         & 32,790 \\
        Total Number of Annotated Frames  & 293,000         & 32,790 \\
        Total Number of Pedestrians       & 1,842           & 1,481  \\
        Crossed Pedestrians               & 519             & 409  \\
        Not Crossed Pedestrians           & 1,323             & 1,072  \\
        \noalign{\hrule height 1pt}
    \end{tabular}
    \label{table:datasets}
    \vspace{-15pt}
\end{table}

\subsubsection{\bf{Urban-PIP dataset}} 
The dataset captures diverse traffic conditions, ranging from busy intersections to quieter residential streets, with dynamic environmental changes. Pedestrian behaviour may be influenced by the actions of nearby vehicles or other pedestrians, making intention prediction more challenging.
It is recorded under various weather conditions, including sunny, cloudy, rainy, and foggy, in three geographical locations using a multi-sensor setup. This multi-modal dataset is collected via a combination of LiDAR, cameras, radar, and IMU sensors mounted on the ego-vehicle. 
LiDAR provides 360$^{\circ}$ field of view with approximately a 300-meter range by beaming out millions of laser pulses per second and measuring the time of laser beam flight from the sensor to the surface of an object, then reflecting from the object to the sensor on the ego vehicle. 
The radar system has a continuous 360$^{\circ}$ view to track the presence and speed of road users in front, behind and sides of the vehicle. 
The multi-camera perspective (left, front, and right) offers a richer contextual understanding of pedestrian interactions compared to single-camera datasets like PIE and JAAD. The cameras simultaneously capture the traffic scene videos in HD format ($1920 \times 1080$),
The IMU module uses accelerometers and gyroscopes with input from GPS, maps, wheel speeds, as well as laser and radar measurements to provide position, velocity, and heading information to the vehicle.

The Urban-PIP dataset captures both daylight and nighttime scenarios, making it more diverse and challenging than datasets such as PIE, JAAD, and STIP, which are limited to daylight conditions. The inclusion of low-light environments introduces additional challenges, where the visibility of pedestrians is reduced, and sensor data may become noisy, leading to challenges in detecting and tracking pedestrians' behaviour accurately.

In this study, the experiments are conducted using camera sensors as they provide rich visual information, including detailed information about pedestrian behaviour, their body language, and contextual information that can be crucial for predicting crossing intentions. Also, the affordability of camera sensors has made them a practical choice for current research on crossing prediction.
We annotated 1,481 pedestrian crossing intentions from the front cameras including 448 in the front-left camera, 541 from the front camera, and 492 from the front-right camera. 

\subsubsection{\bf{Frontal Urban-PIP dataset}} To assess models limited to a single camera, we introduce a subset of the Urban-PIP dataset, named \textit{Frontal Urban-PIP} (FU-PIP), focusing on pedestrians observed only by the front camera. 
This subset, featuring 55 pedestrians with crossing intentions and 129 without, ensures a fair comparison will be conducted with similar methods which are limited to a single camera only.

\subsection{Hardware and Implementation Settings}

All proposed models were executed on a CUDA parallel computing platform with an Nvidia Quadro RTX A5000 GPU, 64GB of RAM and an Intel Core i9 13900K 24-core processor and the Torch environment.

PIP-Net was trained using the RMSProp optimiser. 
256 hidden units were used for the GRUs, and the sigmoid ($\sigma$) activation function was applied to the GRUs for handling spatial kinematic data. To mitigate overfitting, a dropout rate of 0.5 was introduced after the attention block. Additionally, an L2 regularisation term of 0.0001 was incorporated into the last fully connected layer.

\textbf{PIP-Net-$\alpha$}: This model is designed for a single camera setup. The model is trained on the PIE dataset and evaluated against the PIE test set and the FU-PIP datasets. The model doesn't have the Camera Index pipeline within its architecture. Thereby, the outputs of the Max-pooling 3D (MP3D in Figure \ref{fig:method}) are directly forwarded to the flattened block, and the Shift blocks are deactivated.
A learning rate of $5 \times 10^{-5}$ was used for 300 epochs with a batch size 10. The model was tested on the FU-PIP dataset to evaluate its generalisability.

\textbf{PIP-Net-$\beta$}: This model is designed for multi-camera setups to be evaluated against Urban-PIP with three cameras.
The training of this model is performed using the Urban-PIP dataset with various observation times, a learning rate of $4 \times 10^{-5}$ across 400 epochs, and a batch size of 6. 
The split ratio for training and testing samples is 80\% (1,181 samples) and 20\% (296 samples) of the dataset, respectively.

\subsection{Temporal Resolution and Observation Time}
\begin{figure}[!t]
    \centering
    \includegraphics[width=2.4in]{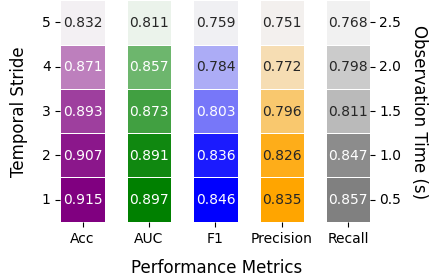}
    \vspace{-2.5mm}
    \caption{Performance of PIP-Net$-\alpha$ across different temporal resolutions: As the stride increases, the sampling frequency decreases, expanding the observation time from 0.5 to 2.5 seconds, influencing the model’s ability to predict pedestrian crossings.}
    \label{fig:temporal_resolution}
    \vspace{-15pt}
\end{figure}

Temporal resolution plays a crucial role in pedestrian crossing prediction, as it determines the frequency at which visual observations are sampled over time. To analyse its impact, we conducted experiments by varying the stride length while keeping the observation sequence length fixed at $m=15$ (processing 15 frames), as followed the benchmark \cite{kotseruba2021benchmark}. This approach allows us to expand the temporal window without increasing the model’s parameters, enabling a trade-off between capturing long-term dependencies and maintaining computational efficiency. We experimented with five different temporal resolutions over PIE videos by varying the stride from 1 (processing every frame) to 5 (processing every fifth frame), leading to observation times ranging from 0.5 seconds to 2.5 seconds.
Figure \ref{fig:temporal_resolution} presents the performance of our proposed model, PIP-Net$-\alpha$, across different temporal resolutions. The results indicate that increasing the stride initially improves performance, as observed with stride 1, which achieves the best results (Acc: 0.915, AUC: 0.897, F1: 0.846). However, beyond stride 2, the performance degrades, suggesting that excessive temporal gaps between frames may lead to loss of fine-grained motion cues critical for precise crossing prediction. This aligns with other studies that suggest the intuition that the closer to crossing events, the more visual information clues will be revealed \cite{Rasouli2020PedestrianAA,zhou2023pit,sharma2023visual}.

\begin{table}[ht]
    \centering
    \caption{Performance results of PIP-Net-$\alpha$, memory usage, inference time, and observation time for different temporal strides and sequence lengths on PIE dataset.
    }
    \vspace{-1.4mm}
    \setlength{\tabcolsep}{5pt} 
    \begin{tabular}{ccccccccc}
        \noalign{\hrule height 1.2pt}
        \textbf{s} & \textbf{m} & \textbf{OT} & \textbf{Params} & \textbf{M} & \textbf{IT} & \textbf{Acc} & \textbf{AUC} & \textbf{F1} \\ 
        \hline
        1 & \multirow{2}{*}{10} & 0.3s & \multirow{2}{*}{5.5M} & \multirow{2}{*}{1.4GB} & \multirow{2}{*}{55ms} & 0.913 & 0.907 & 0.835 \\
        2 &   & 0.5s &  &  &  & \textbf{0.921} & \textbf{0.935} & \textbf{0.882} \\
        \hline
        
        1 & \multirow{2}{*}{15} & 0.5s & \multirow{2}{*}{5.9M} & \multirow{2}{*}{1.8GB} & \multirow{2}{*}{58ms} & \underline{0.915} & 0.897 & 0.846 \\
        2 &   & 1.0s &  &  &  & 0.907 & 0.891 & 0.836 \\
        \hline
        
        1 & \multirow{2}{*}{20} & 0.6s & \multirow{2}{*}{6.6M} & \multirow{2}{*}{2.5GB} & \multirow{2}{*}{62ms} & 0.908 & \underline{0.911} & \underline{0.853} \\
        2 &   & 1.3s &  &  &  & 0.896 & 0.874 & 0.810 \\
        \noalign{\hrule height 1.2pt}
    \end{tabular}
    \label{tab:performance}
    
    {\scriptsize
        \begin{flushleft}
           \textbf{s}: Temporal Stride; \textbf{m}: Sequence Length (frame); \textbf{OT}: Observation Time (second); \textbf{Params}: Number of model parameters (million); \textbf{M}: Memory usage (gigabyte); \textbf{IT}: Inference Time (millisecond).
        \end{flushleft}}
    \vspace{-15pt}
\end{table}

To further investigate the impact of temporal dependencies, we conducted experiments by varying the observation sequence length $m$ while keeping the temporal stride fixed at 1 and 2. The objective was to determine whether increasing the number of frames to process improves prediction performance or introduces redundancy or irrelevant information. 
Table \ref{tab:performance} presents the results, including the number of model parameters, memory usage, and inference time for different configurations. Our findings indicate that a shorter sequence length $(m=10)$ with a temporal stride of 2 $(s=2)$ achieves the highest prediction performance (AUC = 0.938, F1 = 0.882) while maintaining computational efficiency. This configuration $(s=2, m=10)$ keeps the observation time to 0.5 seconds like $(s=1, m=15)$ but with fewer frames processing. Furthermore, this setting requires fewer learnable parameters (5.5M) and lower memory usage (1.4GB), leading to a more efficient inference time of 55ms. These results suggest that increasing sequence length beyond a certain threshold may introduce redundant or irrelevant information about pedestrians' crossing intentions.

\begin{table}[tb]
    \centering
    \caption{Performance comparison on the PIE dataset.}
    \vspace{-1.4mm}
    \begin{tabular}{llccccc}
        \noalign{\hrule height 1pt}
        \multicolumn{1}{c}{\textbf{Model}} & \textbf{Acc} & \textbf{AUC} & \textbf{F1} & \textbf{Precision} & \textbf{Recall} \\ 
        \hline
        ATGC \cite{rasouli2017ICCVW} & 0.59 & 0.55 & 0.39 & 0.33 & 0.47 \\
        Multi-RNN \cite{bhattacharyya2018long}  & 0.83 & 0.80 & 0.71 & 0.69 & 0.73 \\
        SingleRNN \cite{kotseruba2020they}  & 0.81 & 0.75 & 0.64 & 0.67 & 0.61 \\
        SFRNN \cite{Rasouli2020PedestrianAA} & 0.84 & 0.82 & 0.72 & 0.75 & 0.80 \\
        PCPA \cite{kotseruba2021benchmark} & 0.87 & 0.86 & 0.77 & 0.75 & 0.79 \\
        CAPformer \cite{Lorenzo2021CAPformerPC} & 0.88 & 0.80 & 0.71 & 0.69 & 0.74 \\
        PPCI \cite{yang2022predicting}  & 0.89 & 0.86 & 0.80 & 0.79 & 0.81 \\
        GraphPlus \cite{cadena2022pedestrian}  & 0.89 & 0.90 & 0.81 & 0.83 & 0.79 \\
        MCIP$^*$ \hspace{-1mm}\cite{ham2022mcip}  & 0.89 & 0.87 & 0.81 & 0.81 & 0.81 \\
        CIPF$^*$ \hspace{-1mm}\cite{ham2023cipf}  & 0.91 & 0.89 & 0.84 & 0.85 & 0.83 \\ 
        PIT$^*$ \hspace{-1mm} \cite{zhou2023pit} & 0.91 & 0.92 & 0.82 & 0.84 & 0.81 \\
        VMIGI \cite{sharma2023visual} & \textbf{0.92} & 0.91 & 0.87 & 0.86 & \textbf{0.88} \\
        TrEP \cite{zhang2023trep} & \textbf{0.92} & 0.93 & 0.86 & 0.87 & 0.84 \\
        \bf{PIP-Net-$\alpha$ (Ours)} & \textbf{0.92} & \textbf{0.94} & \textbf{0.88} & \textbf{0.89} & \textbf{0.88} \\ 
        \noalign{\hrule height 1pt}
    \end{tabular}
    \label{tab:pie}
    {\scriptsize
    \begin{flushleft}
    \vspace{-1mm}
       \hspace{3mm} All models are evaluated at ETC = 0.5s. Results marked with $^*$ are reported from the original article due to unavailable source code.
    \end{flushleft}}
    \vspace{-15pt}
\end{table}

\subsection{Comparative Results}
Table \ref{tab:pie} highlights the performance of our method on the PIE dataset. The observation time ($m$) has been set to 15 frames, the same as the previous methods, to ensure a fair comparison with other methods. 
PIP-Net-$\alpha$, achieves the highest values in all metrics, showcasing its prowess in capturing the crossing intention classification.

In comparison to MultiRNN, SingleRNN, and SFRNN, which rely on CNN encoders for visual features and RNN-based encoder-decoder structures, PIP-Net-$\alpha$ achieves significant improvements by integrating a more diverse set of input features, capturing both pedestrian kinematics and environmental context more effectively. Unlike transformer-based (CAPformer) and graph-based (GraphPlus) architectures, which struggle with motion generalisation and real-time inference, PIP-Net-$\alpha$ leverages an RNN-based architecture to maintain temporal continuity and robust sequential reasoning, resulting in higher predictive performance. Furthermore, the proposed feature fusion approach, combining three kinematic and four contextual features, enables a more comprehensive scene understanding, improving AUC by +4\% compared to CIPF, which integrates eight pedestrian-vehicle interaction features.

While recent models such as PIT, VMIGI, and TrEP have introduced advanced techniques for long-range dependencies, relational learning, and uncertainty estimation, their effectiveness remains constrained by the inherent limitations of their respective architectures. PIT, incorporating self-attention mechanisms and Temporal Fusion Blocks, enhances sequence modelling but requires heavy computational resources, limiting real-time feasibility. VMIGI, which applies a multimodal fusion strategy with GCN-based relational learning, strengthens interaction reasoning but struggles with fast motion scenarios due to graph structure limitations. TrEP, adopting a transformer-based evidential prediction framework, explicitly models uncertainty by correlating it with human annotator disagreement, yet its reliance on pure transformer encoding leads to weaker sequential motion modelling. In contrast, PIP-Net-$\alpha$ effectively balances multimodal integration, temporal consistency, and computational efficiency, making it the most well-rounded solution for pedestrian crossing prediction in dynamic urban environments, modelling uncertainty and correlating it with human annotator disagreement.

\subsubsection{\bf{Crossing Time Prediction}}
Depending on the traffic scenario, the model's prediction performance may vary. The model can predict the pedestrians' estimated time to cross (ETC), 1 to 4 seconds in advance. For example, an ETC=2 means the model expects or predicts the target pedestrian crosses in 2 seconds.

\begin{figure}[!t]
    \centering
    \includegraphics[width=3.3in]{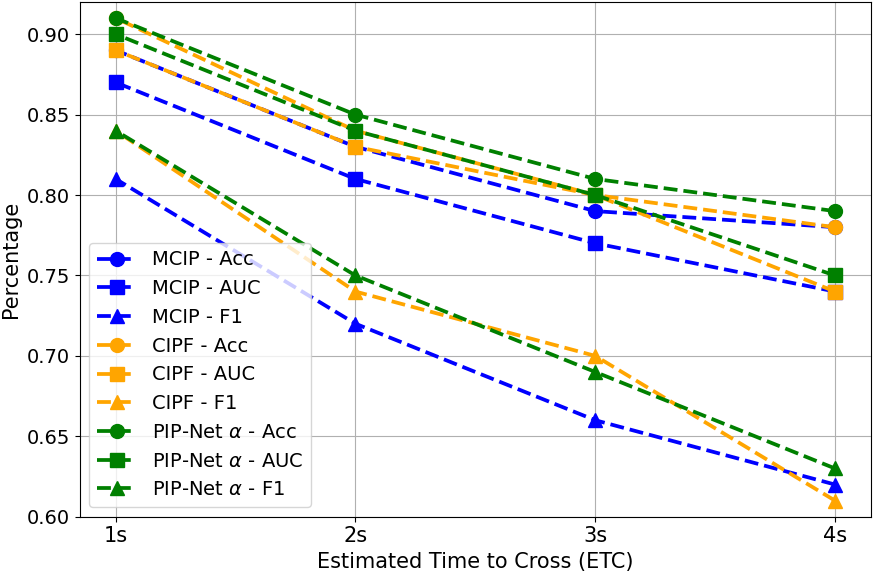}
    \vspace{-3mm}
    \caption{Comparative analysis of prediction performance is conducted for future time points in the study. Models are evaluated in terms of their predictive accuracy from 1 second to 4 seconds ahead.}
    \label{fig:abl_time}
    \vspace{-15pt}
\end{figure}

We evaluated the performance of the proposed PIP-Net-$\alpha$ model across various ETCs from 1 to 4 seconds.
Figure \ref{fig:abl_time} presents a comparison of the Acc, AUC, and F1 performance of PIP-Net-$\alpha$ with two recent prediction models, CIPF and MCIP, using the PIE dataset.
All three models exhibited a decline in performance across all metrics as the Estimated Time to Cross (ETC) increased, i.e. when the models tried to have a longer-term prediction.
Notably, the most significant drop in AUC occurred between the ETC = 1-second and ETC = 2, with MCIP and CIPF models decreasing by 6.8\% and 6.7\%, respectively. This is while our model shows a 6.6\% decrease in the AUC from ETC = 1 to 2. The Accuracy also shows gradual decreases from the 3-second to the 4-second interval, with all models experiencing a decrease of approximately 1\% (MCIP) and 2\% (CIPF), respectively. As can be seen, the proposed model with green dashed lines consistently outperformed the other models for all ETCs.

\subsubsection{\bf{Ablation Study}}
To investigate the inner workings of the proposed models, incorporating interpretability methods such as Grad-CAM \cite{selvaraju2017grad} can enhance understanding by visualising which parts of the image influence predictions. However, in complex models that process heterogeneous input modalities and sequential data, these techniques may not always provide clear or actionable insights. 
Although VIMGI \cite{sharma2023visual} successfully visualised the visual encoder, such methods often focus on individual components, missing feature interactions and kinematic data that affect predictions.
Our approach, similar to studies such as CIPF \cite{ham2023cipf}, MCIP \cite{ham2022mcip}, TrEP \cite{zhang2023trep} and CAPformer \cite{Lorenzo2021CAPformerPC}, emphasises performance-driven modelling, prioritising accuracy and efficiency. 
We have conducted comprehensive ablation studies to address interpretability concerns, providing valuable insights into our model's decision-making processes.

The baseline model, $\alpha^0$, comprises the primary input features recognised in the task of pedestrian intention prediction, including \textit{Bounding Box}, \textit{Body Pose}, \textit{Local Content}, \textit{Semantic Context}, and \textit{Vehicle Speed} \cite{Lorenzo2021CAPformerPC,ham2022mcip,ham2023cipf,ni2023pedestrians,azarmi2024feature}.
Ablation experiments were conducted on the best-performing configuration, as presented in Table \ref{tab:performance}, with s = 2 and m = 10. The results demonstrate that excluding the \textit{Bounding Box} input leads to an 8.6\% decrease in accuracy compared with the baseline, whereas omitting the \textit{Body Pose} parameter results in a lower accuracy reduction of 3.5\%.
This represents less importance of body pose compared to the bounding box which may seem counter-intuitive. However, our further investigations confirm that the bounding box data is notably more important and useful as it includes the pedestrian's moving trajectory and tracking history over time, providing valuable spatio-temporal information. While body poses spatio-temporal information is not very important. The body pose only seems to matter when the pedestrian is about to cross the road in the last few frames before crossing. Otherwise, the body pose in previous moments, such as when the pedestrian is on the sidewalk, is redundant.

The impact of removing \textit{Local Content} causes a decrease in accuracy by 1.7\%. It appears to lack comprehensive features on pedestrians' intentions, given the wide variety in appearance and accessories they may have. 
Excluding \textit{Semantic Context} leads to a 3.4\% accuracy decrease as it includes details about road layout such as sidewalk positioning and drivable zones. 

Interestingly, removing the \textit{Vehicle Speed} feature results in a 3.8\% drop in accuracy, making it the second most important input. This aligns with the findings of the study by \cite{Lorenzo2021CAPformerPC}, which states that a model trained with the ego-vehicle speed tends to focus on the ego-vehicle speed adjustment (e.g. deceleration) to learn the pedestrian intention, rather than learning to predict the intention from the pedestrian behaviour.

\begin{table}[tb]
    \centering
    \caption{PIP-Net-$\alpha$ variants using different input features.}
    \vspace{-1.4mm}
    \setlength{\tabcolsep}{4pt} 
    \begin{tabular}{c|cccc|ccc|c}
        \noalign{\hrule height 1pt}
        \textbf{Variant} & \textbf{GM} & \textbf{LM} & \textbf{MD} & \textbf{CD} & \textbf{Acc} & \textbf{AUC} & \textbf{F1} &  \textbf{IT}\\ 
        \hline
        \textbf{$\alpha^0$} & - & - & - & - &  0.893 & 0.907 & 0.830 & 25\\
        \hdashline
        \textbf{$\alpha^1$} & \checkmark & - & - & - & 0.904$\uparrow$ & 0.913$\uparrow$ & 0.827$\downarrow$ & +10\\
        \textbf{$\alpha^2$} & - & \checkmark & - & - & 0.898$\uparrow$ & 0.919$\uparrow$ & 0.839$\uparrow$ & +3 \\
        \textbf{$\alpha^3$} & - & - & \checkmark & - & 0.884$\downarrow$ & 0.902$\downarrow$ & 0.836$\uparrow$ & +12\\
        \textbf{$\alpha^4$} & - & - & - & \checkmark & \textbf{0.914}$\uparrow$ & \textbf{0.924}$\uparrow$ & \textbf{0.867}$\uparrow$ & +27\\
        \textbf{$\alpha^5$} & \checkmark & - & \checkmark & - & 0.883$\downarrow$ & 0.903$\downarrow$ & 0.827$\downarrow$ & +22\\
        \textbf{$\alpha$} & - & \checkmark & - & \checkmark & \bf{0.921}$\uparrow$ & \bf{0.938}$\uparrow$ & \bf{0.882}$\uparrow$ & +30\\
        \noalign{\hrule height 1pt}
    \end{tabular}
    \label{tab:abl_features}

    {\scriptsize
        \begin{flushleft}
            Features: \textbf{GM} is Global Motion,
            \textbf{LM} is Local Motion,
            \textbf{MD} is  ManyDepth, and 
            \textbf{CD} is Categorical Depth,
            \textbf{IT}: Inference Time (milliseconds).
            The up arrow indicates an improvement, and the down arrow indicates a decrement in comparison with the baseline variant.
        \end{flushleft}}
    \vspace{-15pt}
\end{table}

Table \ref{tab:abl_features} shows a comparison of the two input features used in the proposed model, including scene motion and depth information, along with their impact on inference time.
The $\alpha^2$ variant demonstrates that the \textit{local motion} feature can exhibit superior performance when compared to the global motion feature (\ref{fig:features}d), which includes optical flow analysis of the entire scene. While optical flow is typically sensitive to any movement between consecutive frames, it appears that local motion can provide a coarse-grained feature and treat more concisely to account for pedestrians' velocity and direction of movement, regardless of other irrelevant objects in the surroundings. Additionally, local motion introduces only a modest increase in inference time (+3ms), making it an efficient feature choice.

Regarding depth information features, the categorical depth feature proposed stands out as the most effective standalone feature, as evidenced by the results of variant $\alpha^4$ in Table \ref{tab:abl_features}, highlighting the significance of pedestrians' distance to the ego-vehicle and their interactions with other road users in the traffic scene. However, this improvement comes at the cost of a notable increase in inference time (+27ms).
Conversely, $\alpha^3$ performs the weakest among the variants when compared with the baseline, which utilises the global depth heatmap of the entire scene (\ref{fig:features}e). This underperformance may be attributed to the unstable estimation of depth for irrelevant surrounding objects. This is addressed in the categorical depth by focusing only on pedestrians and vehicles, and then applying per-instance normalisation (as shown in Figure \ref{fig:categorical_depth}).  

Finally, the optimal outcome is attained by taking into account both the local motion and categorical depth map in $\alpha$, leading to enhancements in the baseline regarding Acc, AUC, and F1 score by 3.14\%, 3.1\%, and 6.2\%, respectively. This improvement, however, comes with the highest inference time (+30ms), emphasising the trade-off between model complexity and real-time applicability. The full model ($\alpha$) achieves 55 milliseconds ($\approx$ 18 FPS) inference time while maintaining high prediction performance, ensuring its suitability for real-time autonomous systems.

\subsubsection{\bf{Generalisation}}
The PIP-Net-$\alpha$ evaluation is presented against the SOTA methodologies on the FU-PIP dataset in Table \ref{tab:Urban-PIP-a}. Notably, the models have never seen the scenarios in their training phase. The outcomes of other methods were generated using the pre-trained weights they provided.
Overall, models exhibit limited generalisation capability on Urban-PIP, suggesting that our dataset is a more challenging benchmark compared to PIE. Given its environmental variety and inclusion of nighttime scenes, we believe our findings are largely transferable to PIE and JAAD, which mainly comprise daylight conditions and may not fully capture real-world complexities.

\begin{figure}[!t]
    \centering
    \includegraphics[width=3.2in]{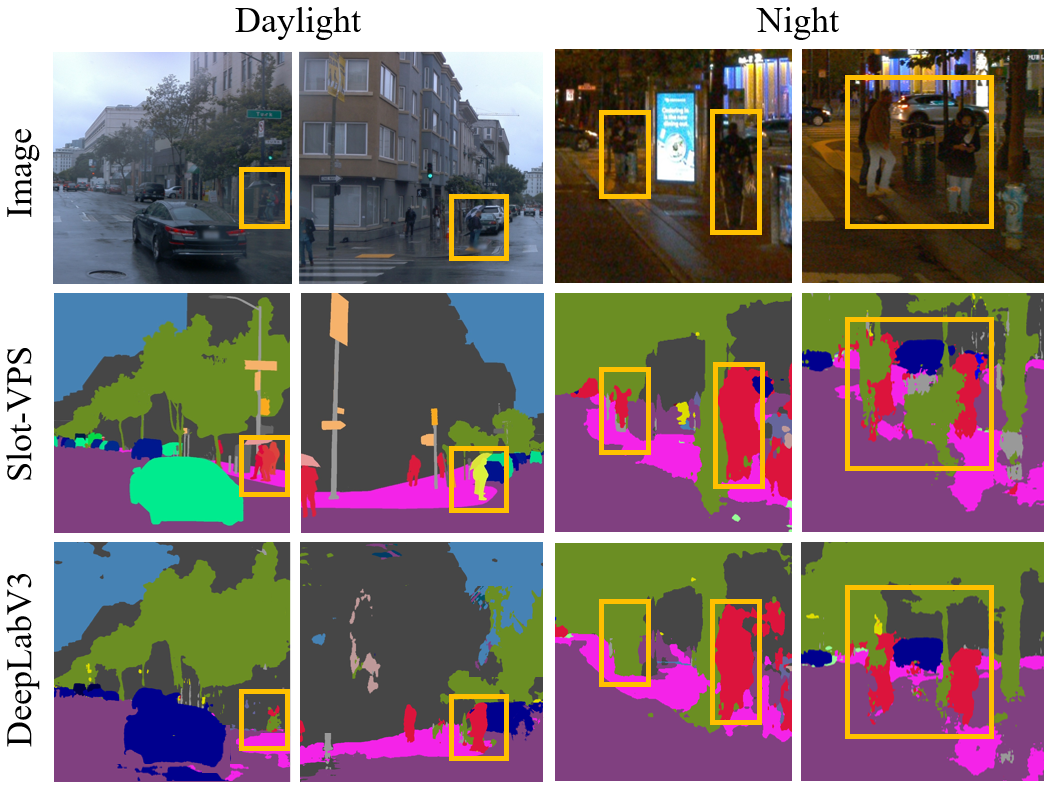}
    \vspace{-9pt}
    \caption{Comparison of semantic and panoptic segmentation, where panoptic segmentation offers more precise scenes in daylight by distinguishing object instances and semantic categories, leading to improved global context representation.}
    \label{fig:pano}
    \vspace{-10pt}
\end{figure}

Most models show improvements over ATGC across various metrics, with each demonstrating unique strengths. 
However, we witnessed a decrement in terms of performance for PCPA, PPCI, CAPformer, and GraphPlus. As far as the research curiosity demand, we investigated the architecture of these models, and it turned out they are suffering from low-quality global context (see Figure \ref{fig:pano}) and body pose features. This is caused by the feature extractor algorithms, i.e., the semantic segmentation and pose estimator they have used, which hinder the classifier from judging based on precise features.

In our model, we enhanced semantic context representation by employing panoptic segmentation instead of conventional semantic segmentation. Unlike semantic segmentation, which classifies each pixel into a category for a single image, panoptic segmentation processes video sequences to generate temporally consistent object segments, distinguishing between object instances and semantic categories simultaneously. 
However, failure cases arise in challenging scenarios, particularly in low-light conditions or occluded scenes.
Figure \ref{fig:pano} illustrates qualitative comparisons between Slot-VPS (used in our model) and DeepLabV3 \cite{chen2017rethinking} (used in previous works). During daytime, Slot-VPS \cite{zhou2022slot} provides sharper instance boundaries and more stable object tracking. However, at night, we observe increased misclassification and temporal inconsistencies, where pedestrians may be incorrectly segmented as part of the background or falsely merged with nearby objects. These segmentation errors propagate to downstream predictions, potentially leading to incorrect pedestrian intention estimates, especially when multiple pedestrians are present.

\begin{table}[tb]
    \centering
    \caption{Performance comparison on the Frontal Urban-PIP dataset.}
    \begin{tabular}{lccccc}
        \noalign{\hrule height 1pt}
        \multicolumn{1}{l}{\textbf{Model}} & \textbf{Acc} & \textbf{AUC} & \textbf{F1} & \textbf{Precision} & \textbf{Recall} \\ 
        \hline
        ATGC      & 0.52 & 0.51 & 0.35 & 0.32 & 0.44 \\
        Multi-RNN & 0.64 & 0.63 & 0.49 & 0.51 & 0.48 \\
        SingleRNN & 0.65 & 0.64 & 0.54 & 0.57 & 0.53 \\
        SFRNN     & 0.65 & 0.65 & 0.55 & 0.58 & 0.53 \\
        PCPA      & 0.62 & 0.60 & 0.50 & 0.52 & 0.47 \\
        PPCI      & 0.63 & 0.61 & 0.51 & 0.55 & 0.47 \\
        CAPformer & 0.64 & 0.60 & 0.55 & 0.58 & 0.54 \\
        GraphPlus & 0.64 & 0.61 & 0.57 & 0.59 & 0.56 \\
        VMIGI     & 0.65 & 0.61 & 0.57 & 0.58 & 0.55 \\
        TrEP      & 0.64 & 0.62 & 0.58 & 0.59 & 0.57 \\
        \bf{PIP-Net-$\alpha$} (Ours) & \bf{0.73} & \bf{0.71} & \bf{0.69} & \bf{0.70} & \bf{0.68} \\ 
        \noalign{\hrule height 1pt}
    \end{tabular}
    \label{tab:Urban-PIP-a}
\end{table}

\begin{figure}[!t]
    \centering
    \includegraphics[width=2.8in]{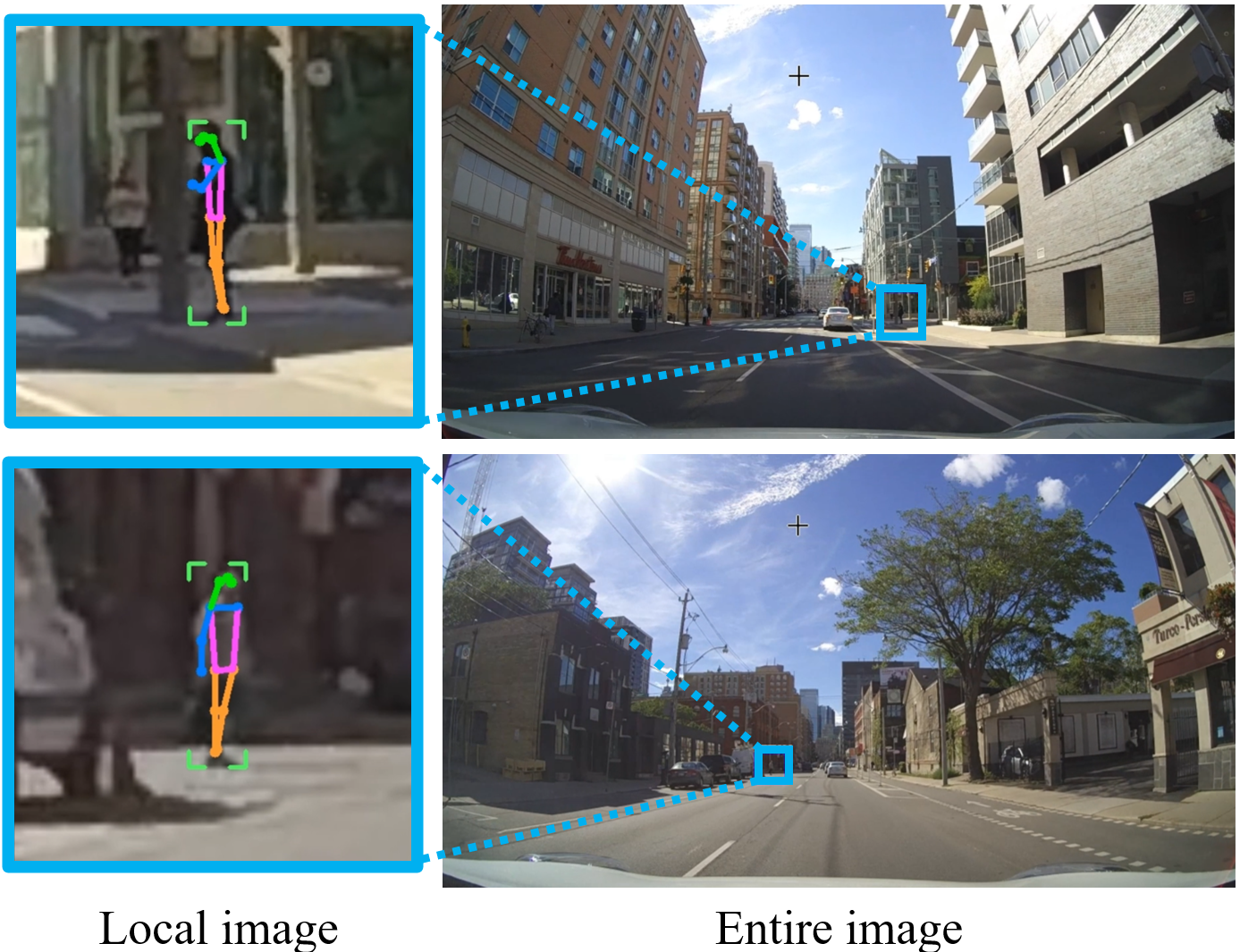}
    \vspace{-8pt}
    \caption{Body pose estimation on local image rather than on the entire image enables more precise and reliable pose extraction.}
    \label{fig:look}
    \vspace{-15pt}
\end{figure}

We observed more precise body pose information by applying pose estimation on the local image region containing the pedestrian rather than the entire scene image. This method ensures a more focused and accurate pose representation, reducing noise and ambiguity caused by irrelevant background elements. Figure \ref{fig:look} illustrates how our approach successfully retrieves clear and precise pose data in scenarios where applying pose estimation on the entire scene fails to obtain reliable pose information. Applying the pose estimation algorithm to the local image region results in more robust and consistent pose features.

While these enhancements contribute to better generalisation, performance variations in challenging conditions emphasise the need for further research into feature extraction under adverse lighting and occlusions to ensure model reliability across all operational scenarios.

\subsubsection{\bf{View Angle Expansion}}
\begin{figure}[!t]
    \centering
    \includegraphics[width=3.5in]{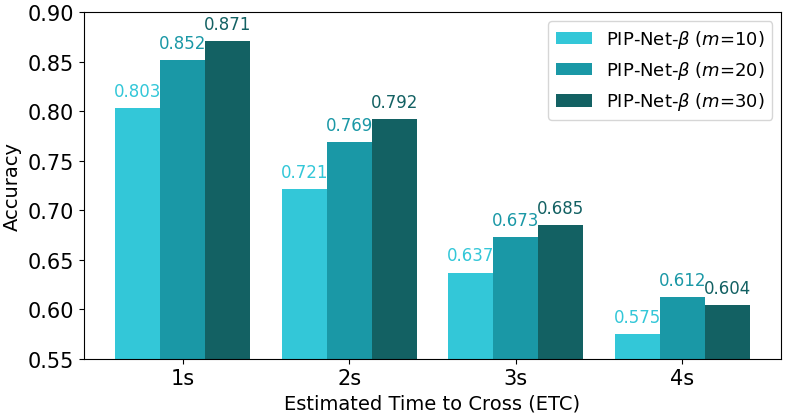}
    \vspace{-15pt}
    \caption{Comparison of the crossing intention accuracy (Acc) for PIP-Net-$\beta$ models (in three cameras mode)  across various observation sequence lengths ($m$). The results are based on the Urban-PIP dataset.}
    \label{fig:abl_time_angle}
    \vspace{-12pt}
\end{figure}

We explored the enhancement of the field of view using three cameras to enable the autonomous vehicle to perceive a larger portion of its surroundings. For this purpose, we train PIP-Net-$\beta$ with three different observation times ($m$) of 10, 20, and 30 frames. Subsequently, we examined how the prediction performance evolves as expand the ETC prediction horizon from 1 to 4 seconds.
As depicted in Figure \ref{fig:abl_time_angle}, the accuracy of crossing intention prediction decreases as the ETC prediction expands.
However, the accuracy often increases with the enlargement of observation time. Intriguingly, when ETC = 4 and $m=30$ the accuracy was lower compared to $m=20$.
This discrepancy arises from the model predicting pedestrians to be crossing based on long-term observations when, in reality, they did not cross. This highlights the observation that a pedestrian's previous actions do not always accurately indicate their future intentions, as they can change their mind and act in an instant \cite{najmi2023human}.

\subsubsection{\bf{Computational Cost of Multi-Camera Integration}}
To assess the computational cost of integrating three cameras, we evaluated the inference time, memory usage, and parameter count of PIP-Net-$\beta$ with stitched multi-camera inputs. Our results indicate that while computational demands increase, the model remains suitable for real-time execution. Specifically, the three-camera version required 9.9M parameters, 2.9GB of memory, and an inference time of 127ms ($\approx$ 7.8 FPS). This represents an approximate 1.8× increase in model size, 2.1× in memory usage, and 2.3× in inference time compared to the single-camera setup (PIP-Net-$\alpha$). The feature integration module efficiently mitigates redundancy, ensuring that multi-camera integration remains feasible for real-world deployment.

\subsection{Observational Results}
\begin{figure}[!t]
    \centering
    \subfloat[A non-signalised junction ]{\includegraphics[width=3.3in]{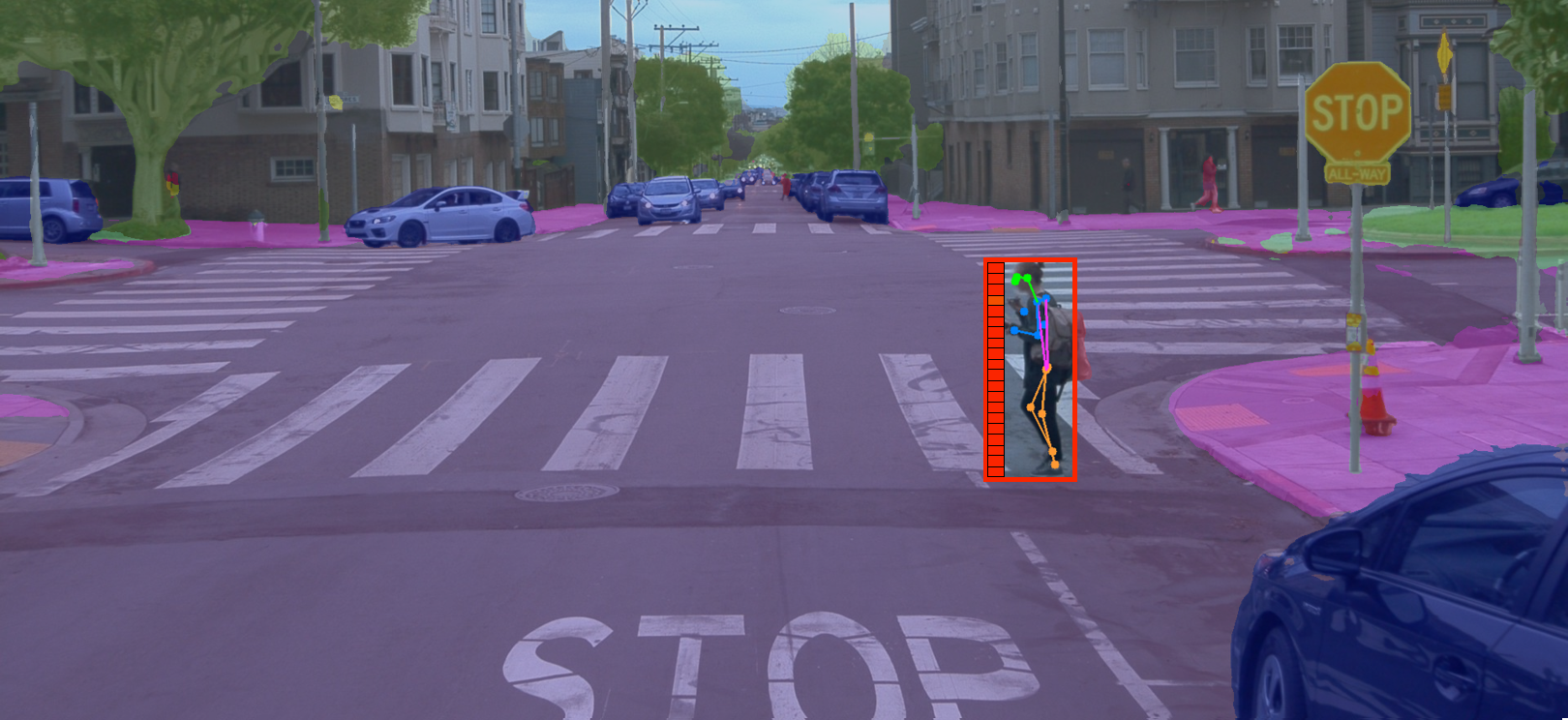}\label{subfig:non-sig}}
    \hfil
    \vspace{-2mm}
    \subfloat[A signalised junction with a blue light]{\includegraphics[width=3.3in]{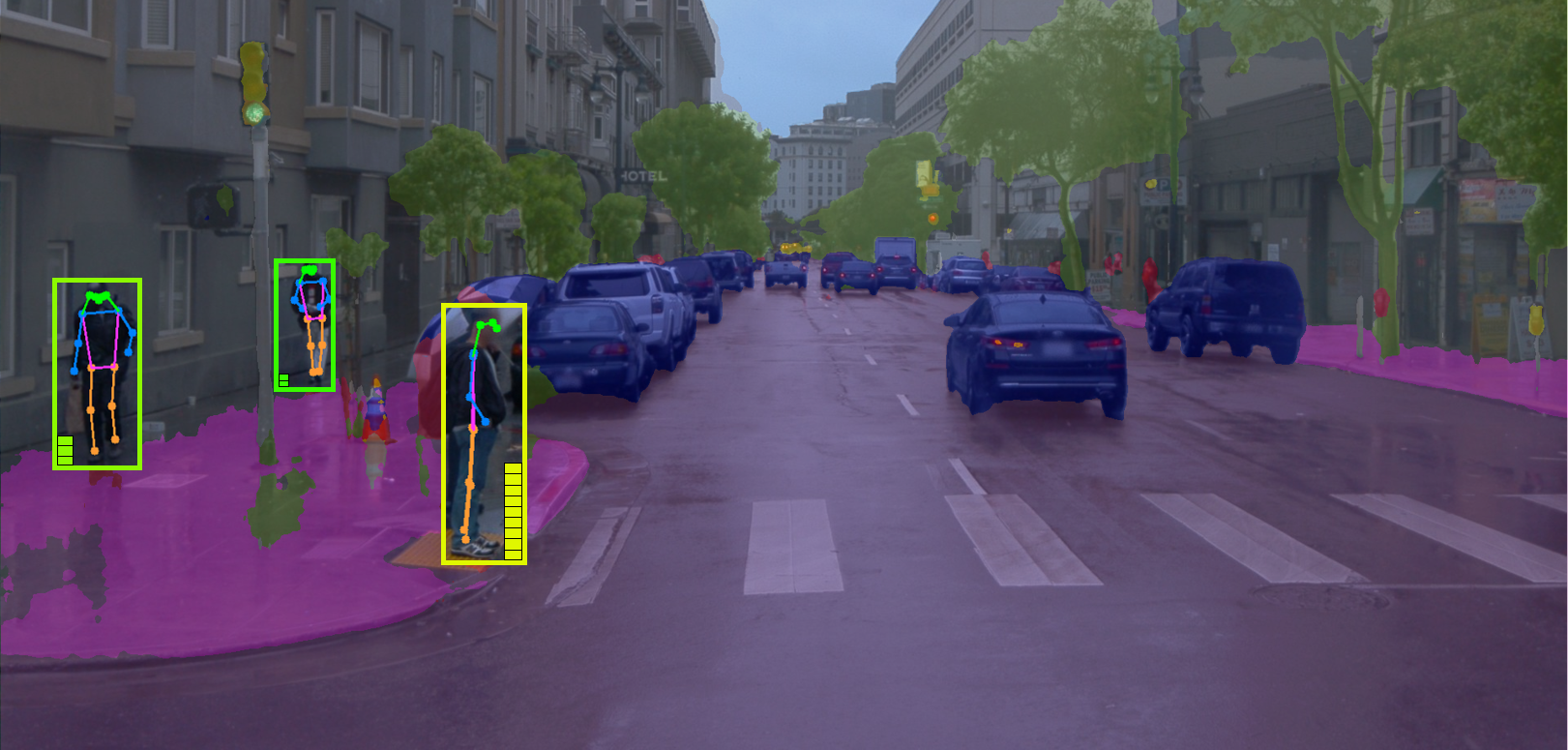}\label{subfig:sig-blue}}
    \hfil
    \vspace{-2mm}
    \subfloat[A signalised junction with a red light]{\includegraphics[width=3.3in]{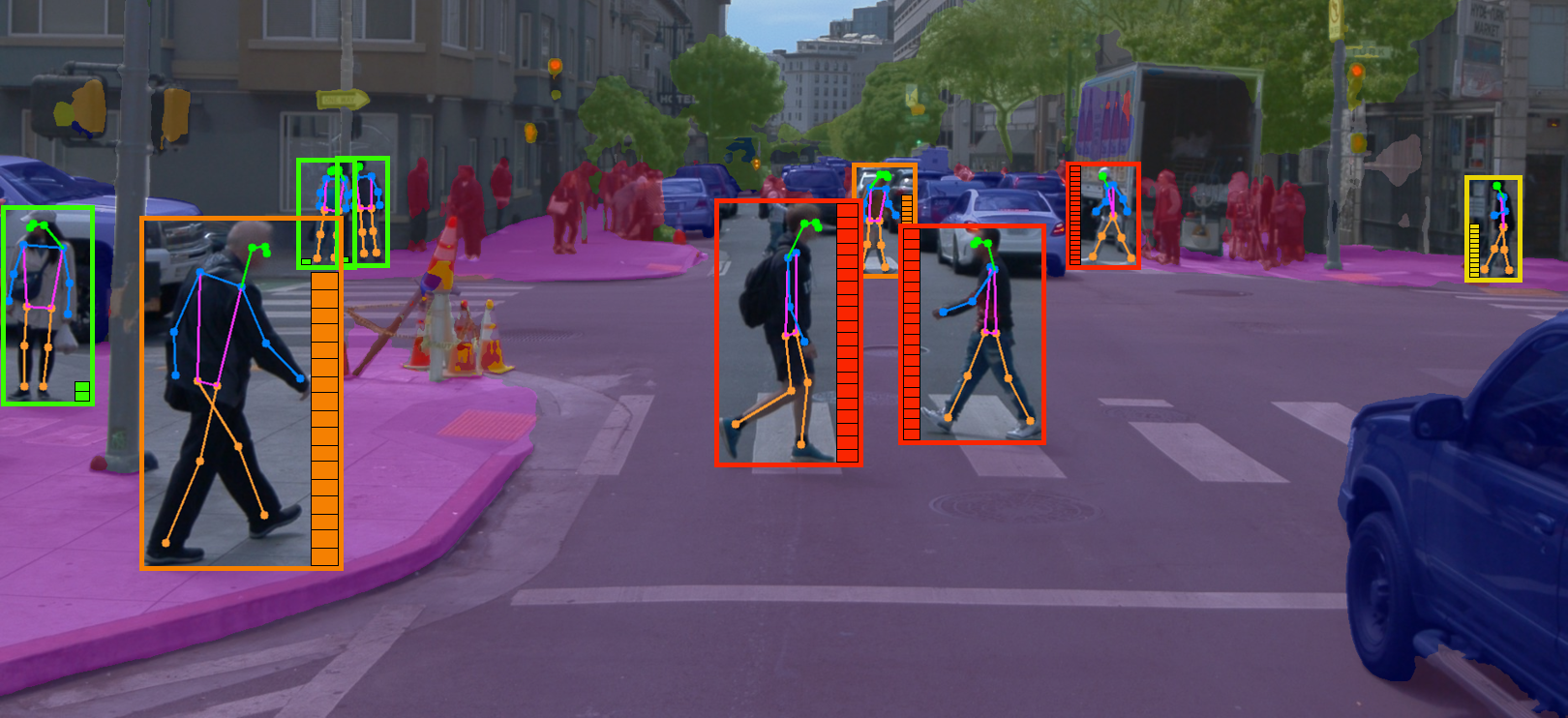}\label{subfig:sig-red}}
    \hfil
    \caption{\textbf{A representation of the proposed model's analysis on traffic video scenes of FU-PIP dataset.} Three distinct images showcase intersections and traffic light conditions with probability bars positioned in front of pedestrians, aligning with their body pose direction. These bars visually represent the model's predicted probability of each detected pedestrian crossing within the next 2 to 4 seconds into the future.}
    \label{fig:results_Urban-PIP}
    \vspace{-15pt}
\end{figure}

We present the qualitative results of the PIP-Net-$\alpha$ in Figure \ref{fig:results_Urban-PIP} for FU-PIP datasets. The crossing probability is represented by a probability bar, where higher values (reddish colour) indicate a high probability of the pedestrian crossing. Pedestrians with no intention to cross are depicted with greenish bounding boxes and lower values on the probability bar.

\begin{figure}[!t]
    \centering
    \includegraphics[width=3.5in]{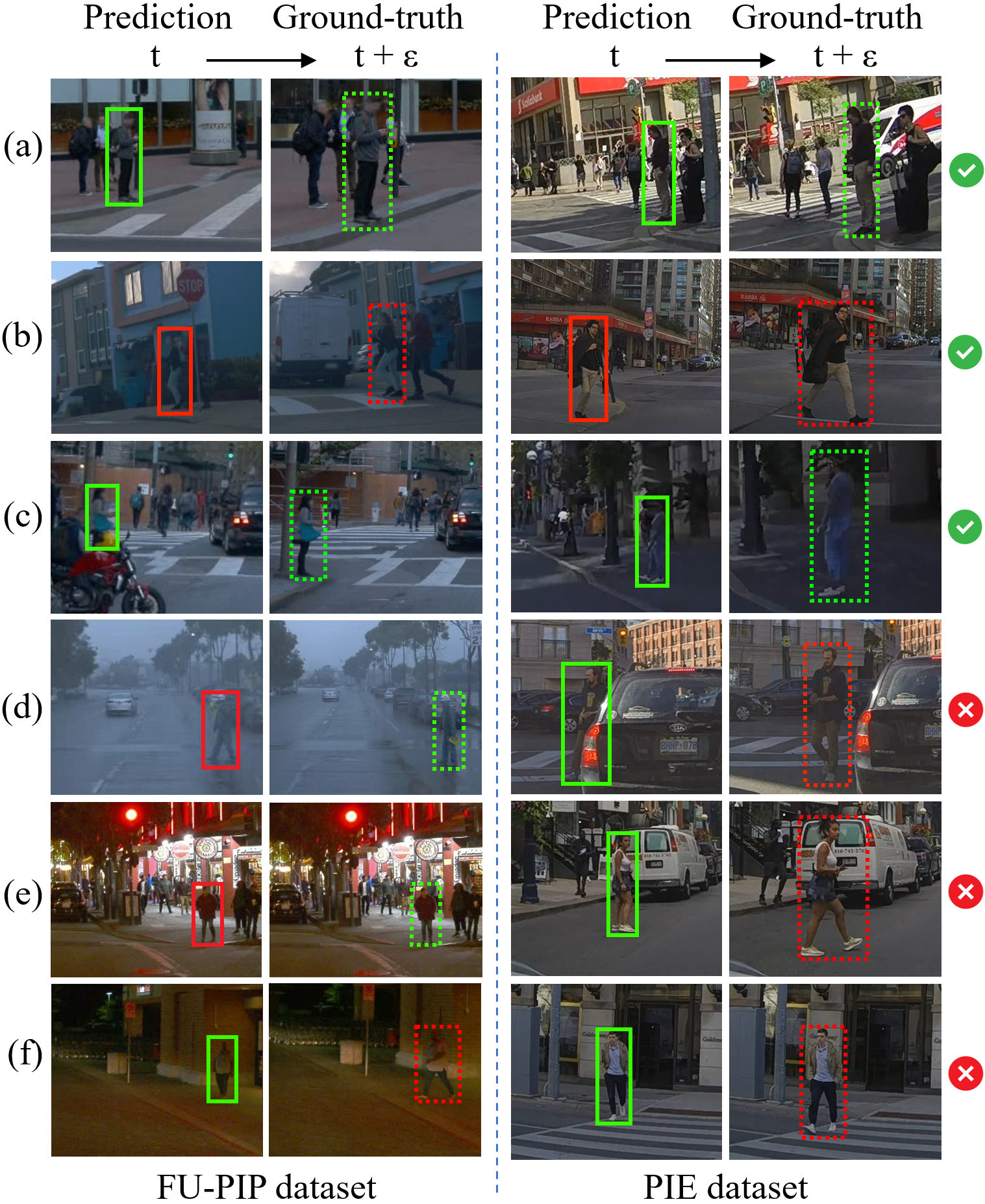}
    \vspace{-5mm}
    \caption{Qualitative results on 12 sample scenarios. Rows with blue ticks represent successful predictions by the model and red crosses indicate wrong predictions. $t$ demonstrates the prediction at the decisive moment and ($t+\epsilon$) shows the ground truth (dashed bounding boxes).  
    Red bounding boxes denote crossing, while blue denotes not crossing.}
    \label{fig:results}
    \vspace{-15pt}
\end{figure}

Figure \ref{fig:results} presents the qualitative results of PIP-Net-$\alpha$ on 12 sample scenarios in the PIE and FU-PIP dataset. 
The model successfully predicted cases in rows (a) to (c), demonstrating robustness across varying daylight ambient lighting conditions. 
In row (c) of FU-PIP, the model accurately inferred the crossing intention of a pedestrian despite partial occlusion. However, in row (d) of PIE, the model struggled to make a correct prediction, when the pedestrian was nearly fully occluded.

In row (d) of FU-PIP, it seemed the model correctly predicted the intention of the pedestrian who stepped onto the road for crossing. However, the pedestrian subsequently abandoned the crossing and stepped back. In such rare cases, where the pedestrian ultimately does not cross, the ground truth is labelled as “not crossing”, consistent with the annotation criteria of the PIE and JAAD datasets.

In row (e) of PIE, the ego vehicle was accelerating, and a hesitation in crossing was observed in the pedestrian's behaviour. Interestingly, the model predicted a 43\% probability of crossing, but this is below the 50\% probability threshold to be accepted as a crossing event. However, we hypothesised that the acceleration of the ego vehicle might have a dominant effect over visual observations. To investigate this, we applied the permutation feature importance approach, as suggested in \cite{azarmi2024feature}, and replaced the speed value with a scenario in which the ego vehicle was decelerating. Notably, the model then predicted an 88\% probability of crossing intention for this case. This finding aligns with previous studies suggesting that, in certain scenarios, the model prediction heavily relies on ego-vehicle speed.

In row (f) of PIE, the ego vehicle was preparing to turn left while the pedestrian was inclined to cross the intersecting road. This suggests that incorporating additional motion-related information about the ego vehicle, such as the heading angle/steering angle, could help to address such cases.

We also evaluated the model’s performance in nighttime scenarios, as shown in rows (e) and (f) of FU-PIP. In these cases, the model failed to classify pedestrian crossing intentions accurately. This poor performance can be attributed to two primary factors: (1) the reduced accuracy of input features extracted from panoptic segmentation and depth estimation under low-light conditions, and (2) the limited number of nighttime samples in the training dataset, which likely led to model underfitting in such scenarios.

\section{Conclusion and Future Work}
This paper presented a framework called PIP-Net for predicting pedestrian crossing intentions in real-world urban self-driving situations. Two variants, $\alpha$ and $\beta$, were introduced to support different camera setups.
By utilising both kinematic data and spatial features of the driving scene, the proposed model employed a recurrent and temporal attention-based methodology to predict pedestrians' future crossing intentions accurately. Through quantitative and qualitative experiments on the PIE dataset, the proposed model achieved state-of-the-art performance with a 94\% AUC and an 88\% F1 score.

The Urban-PIP was introduced as a new dataset for the pedestrian crossing prediction task, including various AV driving scenarios and comprehensive annotations on a multi-sensory setup, thereby enabling a better future investigation of crossing behaviour studies.
Our model demonstrated a generalisation capability when applied to the Urban-PIP dataset by +9\%, +10\%, and 12\% improvement compared with other models in terms of accuracy, AUC, and F1 score, respectively. This was underlined by the scene feature extractors employed in training our model.

To enhance the visual encoding of road users and their relative distances to the ego vehicle, we introduced a categorical depth feature map. This, combined with the sliced motion flow feature, provided salient information about the dynamics of the scene. Our results reveal that they cumulatively enhanced the accuracy and F1 score of the baseline model by +3.1\% and +6.2\%, respectively.
Additionally, we investigated the impact of expanding the view angle using three cameras and enlarging prior observation frames.

Our algorithm achieved 85.4\% accuracy in predicting pedestrian crossing intentions 2 seconds in advance and 79.3\% accuracy for predictions between 2 and 4 seconds in advance.
However, the algorithm is sensitive to the quality and precision of the input features, specifically, scene context and body pose information.

Overall, this study paved the way for developing multi-modal solutions for pedestrian crossing prediction
and provided insights to effectively protect vulnerable road users by foreseeing the crossing behaviour of nearby pedestrians. Following the approach taken by seminal works in the field this study also focused on RGB cameras; however, we extended the use of a single camera to multiple cameras (left, front, and right) to evaluate the effectiveness of contextual information expansion. As one of the possible future work, we recognise the potential benefits of integrating data from other sensors, such as LiDAR and radar, to further enhance the model robustness in adverse weather or challenging lighting conditions. Specifically, LiDAR can offer more precise and reliable depth estimation, enhancing the Categorical Depth representation beyond what can be achieved with monocular-based depth estimation algorithms.

\section*{Declaration of competing interest}
The authors declare that they have no known competing financial interests or personal relationships that could have appeared to influence the work reported in this paper.

\section*{Acknowledgements}
The authors would like to thank all partners within the Hi-Drive project for their cooperation and valuable contribution. This research has received funding from the European Union’s Horizon 2020 research and innovation programme, under grant Agreement No 101006664. The article reflects only the author’s view based on limited datasets and experiments, and neither the European Commission nor CINEA is responsible for any real-world use that may be made of the information this document contains.

\bibliographystyle{IEEEtran}
\bibliography{ref.bib}

\null

\section{Biography}

\vspace{-30pt}
\begin{IEEEbiography}[{\includegraphics[width=1in,height=1.25in,clip,keepaspectratio]{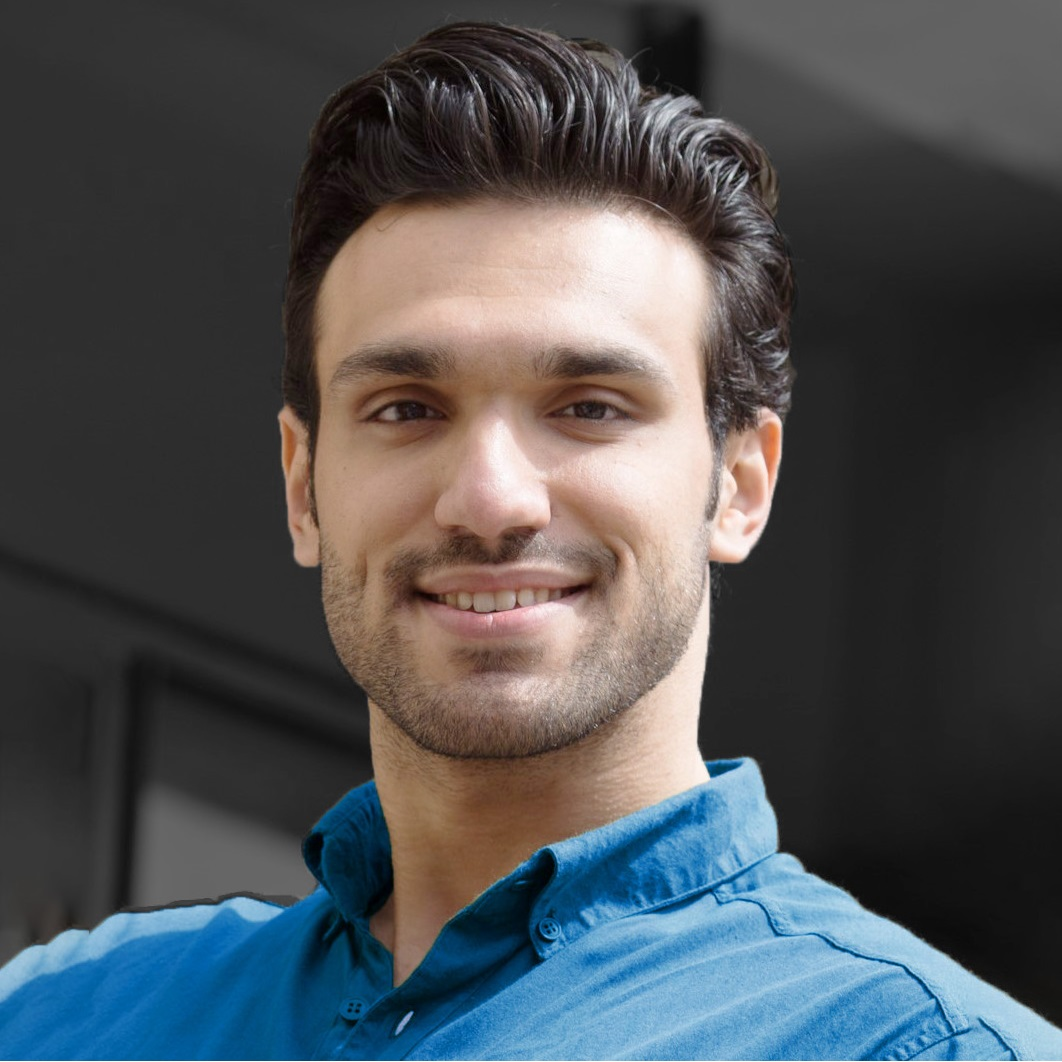}}]{Mohsen Azarmi}
is a Ph.D. student at the Institute for Transport Studies, University of Leeds, UK. He holds a master’s degree in Artificial Intelligence and Robotics. His research focuses on the intersection of computer vision, deep neural networks, and multi-sensor data fusion, with particular interests in pedestrian behaviour understanding, intelligent transportation systems, traffic safety, 3D scene modelling, and multimodal large language models.  
\end{IEEEbiography}

\vspace{-30pt}
\begin{IEEEbiography}[{\includegraphics[width=1in,height=1.25in,clip,keepaspectratio]{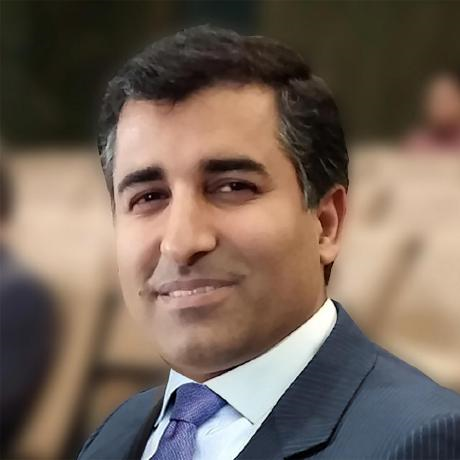}}]{Mahdi Rezaei}
is an Associate Professor in Computer Vision and Machine Learning and the team leader of the Computer Vision Research Group at the University of Leeds, Institute for Transport Studies. With a background in Artificial Intelligence and 18 years of research experience in academia and industry, he is a PI and lead CO-I in various EU and UK flagship projects, such as EPSRC DTP, Hi-Drive, MAVIS, and L3Pilot, with a focus on Driver Monitoring Systems, AV-Pedestrian interactions, XAI, and Intelligent Transportation Systems. 
\end{IEEEbiography}

\vspace{-23pt}
\begin{IEEEbiography}[{\includegraphics[width=1in,height=1.25in,clip,keepaspectratio]{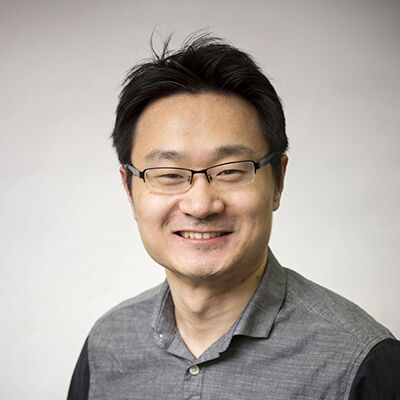}}]{He Wang}
is an Associate Professor at the Department of Computer Science, University College London (UCL) and a Visiting Professor at the University of Leeds. He serves as an Associate Editor of Computer Graphics Forum, an Academic Advisor at the Commonwealth Scholarship Council, and a former Fellow of the Alan Turing Institute.  His current research interest is mainly in computer graphics, computer vision and machine learning. 
\end{IEEEbiography}

\vfill

\end{document}